\documentclass{article}

\usepackage{microtype}
\usepackage{graphicx}
\usepackage{subcaption}
\usepackage{booktabs} 

\usepackage[dvipsnames]{xcolor}

\usepackage{dsfont}  
\usepackage{upgreek} 
\usepackage{mathtools} 

\usepackage{amsthm}
\newtheorem{example}{Example}

\newcommand{\mcal}[1]{\mathcal{#1}}
\newcommand{\mact}[1]{\breve{\mathds{#1}}}
\newcommand{\ffwwitok}{({\color{blue}(\mathcal{W}_i, w_i)})_{i=1}^k}
\newcommand{\ffwitok}{{\color{blue}(\mathcal{W}_i)}_{i=1}^k}
\newcommand{\ffwi}{{\color{blue}\mathcal{W}_i}}

\newcommand{\ffwilitok}{{\color{blue}(\mathcal{W}_i^l)}_{i=1}^k}

\newcommand{\ffwl}{{\color{blue}\mathcal{W}^l}}
\newcommand{\ffwiprevitok}{{\color{blue}(\mathcal{W}_i^{\rm prev})}_{i=1}^{k_{\rm prev}}}

\newcommand{\ffwprev}{{\color{blue}\mathcal{W}^{\rm prev}}}

\newcommand{\orthopitok}{{\color{BlueViolet}(P_i)}_{i=1}^k}
\newcommand{\orthopi}{{\color{BlueViolet}P_i}}
\newcommand{\projectui}{{\color{cyan}U_i}}

\newcommand{\nint}[1]{\lfloor #1\rceil} 
\usepackage{color, colortbl}
\definecolor{LightCyan}{rgb}{0.88,1,1}
\definecolor{Gray}{gray}{0.9}
\definecolor{cgreen}{rgb}{0.38, 0.85, 0.21}

\usepackage{tabularx}
\usepackage{algorithm}
\usepackage{bm}
\usepackage{paralist}

\usepackage{hyperref}

\usepackage[accepted]{icml2024}

\usepackage{amsmath}
\usepackage{amssymb}
\usepackage{mathtools}
\usepackage{amsthm}

\usepackage[capitalize,noabbrev]{cleveref}

\theoremstyle{plain}
\newtheorem{theorem}{Theorem}[section]

\theoremstyle{definition}
\newtheorem{definition}[theorem]{Definition}

\theoremstyle{remark}

\usepackage[textsize=tiny]{todonotes}

\icmltitlerunning{Flexible Low-Bit Quantization for Transformers}

\begin{document}

\twocolumn[
\icmltitle{FrameQuant: Flexible Low-Bit Quantization for Transformers}



\icmlsetsymbol{equal}{*}

\begin{icmlauthorlist}
\icmlauthor{Harshavardhan Adepu}{uwm}
\icmlauthor{Zhanpeng Zeng}{uwm}
\icmlauthor{Li Zhang}{comp}
\icmlauthor{Vikas Singh}{uwm,comp}
\end{icmlauthorlist}

\icmlaffiliation{uwm}{University of Wisconsin-Madison}
\icmlaffiliation{comp}{Google Research}

\icmlcorrespondingauthor{Harshavardhan Adepu}{adepu@wisc.edu}

\icmlkeywords{Machine Learning, ICML}

\vskip 0.3in
]



\printAffiliationsAndNotice{}  

\setcounter{MaxMatrixCols}{20}
\begin{abstract}
Transformers are the backbone of powerful foundation models for many Vision and Natural Language Processing
tasks. But their compute and memory/storage footprint is large, and so, serving such models is expensive often requiring high-end hardware. To mitigate this difficulty, 
Post-Training Quantization seeks to modify a pre-trained 
model and quantize it to eight bits or lower, significantly boosting compute/memory/latency efficiency. Such models have been successfully quantized to four bits with some performance 
loss. In this work, we outline a simple scheme to quantize Transformer-based models
to just two bits (plus some overhead) with only a small drop in accuracy. Key to our formulation 
is a concept borrowed from Harmonic analysis called Fusion Frames. Our main 
finding is that the quantization must take place not 
in the original weight space, but instead in the
Fusion Frame representations. 
If quantization is interpreted as the addition of noise, our 
casting of the problem allows invoking an extensive body of 
known consistent recovery and noise robustness guarantees. Further, 
if desired, de-noising filters are known in closed form. We show empirically, via a variety 
of experiments, that (almost) two-bit quantization for Transformer models promises sizable efficiency gains. The code is available at \url{https://github.com/vsingh-group/FrameQuant}

\end{abstract}
\section{Introduction}
Transformer-based Large Language Models (LLMs) dominate the landscape for Natural Language Processing tasks such as language translation and text summarization \cite{10.5555/3618408.3620130, touvron2023llama, zhang2022opt}. Vision Transformers (VITs) adapt this idea for computer vision, and achieve state-of-the-art results on image classification \cite{zhai2022scaling}, object detection \cite{zhang2022dino}, generation \cite{Chang_2022_CVPR, pmlr-v139-hudson21a} and segmentation \cite{cheng2022masked, ranftl2021vision}. There is general agreement that scale provides remarkable new capabilities.

\comment{
Transformer based large language models (LLMs) dominate the landscape  
for natural languages processing tasks such as language translation and text summarization \cite{}.
Vision Transformers (VITs) adapt this idea for computer vision, and achieve state-of-the-art results on image classification \cite{}, object detection \cite{}, generation \cite{see-https://arxiv.org/pdf/2203.09795.pdf} and segmentation \cite{}. 
Driven in part by the excitement surrounding LLMs, language models have grown in size quickly. 
There is general agreement that scale provides remarkable new 
capabilities. Several language models available today are more than an order 
of magnitude larger than the biggest Vision transformer models. 
Efforts to scale up vision transformers seek to close this gap, and
are based on the premise 
that we will obtain similar ``leveling-up'' of capabilities and performance across the board. Recent results shows that this assumption does appear to hold \cite{vit-22b}.
}

While large models offer strong performance improvements, 
their deployment as a module within a product creates 
unique challenges. For example, serving these models 
on expensive hardware can drastically increase data center 
costs. Even loading these models on consumer-grade machines is difficult, and the ability to handle heterogeneous resource-constrained devices is almost infeasible. This has led to various efficiency-focused strategies for 
model compression including but not limited to 
distillation \cite{hinton2015distilling, zhu2021data}, pruning \cite{8485719}, sparsity \cite{6288897, yun2020n} 
and quantization \cite{DBLP:journals/corr/HanMD15, banner2019post}. Among these methods, Post-Training Quantization offers unique advantages in that it does not change the model architecture or training scheme. 

\comment{
Those who prefer quantization appreciate that the 
method does not change the 
network architecture at all and instead reduces the 
resolution/precision at which the weights (model parameters) are represented -- for instance, 
eight or even four bits. Quantization during training is possible \cite{}
but may require special strategies to avoid numerical instabilities. 
An alternative, that avoids re-training, is post-training quantization (PTQ) \cite{}. 
Essentially, a pre-trained model is {\em post}-quantized to the desired 
number of bits such that the overall functionality of 
the unquantized model is preserved to the extent possible. Such 
a scheme was initially shown to yield excellent performance 
for models like Resnet50 and DeeplabV3 \cite{up-or-down}, but has since 
been shown to work well also for a broad range of 
vision transformers, as well as transformer-based 
language models with tens of billions of parameters \cite{gptq}.
}

\comment{
Transformers emerge as an effective architecture for deep learning models for tasks such as Natural Language Processing, Image Classification, Object detection etc., However, because of their huge size and computational requirements, running these models on compute-limited devices such as edge devices poses numerous challenges. Model compression via Quantization has been effective in reducing the sizes of these models. There are several strategies that have been employed in the past for effective model compression such as quantization-aware training and post-training quantization. Although quantization-aware \cite{from up or down} training is effective, it requires re-training the the models and hence requires significant amount of resources to quantize a model. \\
On the other hand, post-training quantization has become an effective method for quantizing models that only require a small amount of calibration dataset with no training at all. Several works have shown this to be an effective method for quantizing large models to ultra-low bit widths with minimal loss in performance \cite{UrD, OBQ, GPTQ, QUIP, some of nagel's papers}. \textcolor{blue}{Talk more about post training quantization.}\\
}


{\bf This paper} presents a new Post-Training Quantization scheme, FrameQuant, that offers much more flexibility to strike a balance between reducing model size and preserving model quality. Specifically, FrameQuant offers what may be considered equivalent to using a fractional number of bits for quantization, e.g., 2.1 or 2.2 bits: this is valuable because for large Transformer-based models like GPT, model quality deteriorates fast \cite{frantar2023optq} as we reduce bit width in the low-bit quantization regime (e.g., 2-bit). Further, depending on the accuracy needs of the downstream task at hand or a desire to control the worst-off error, more flexibility offers   
the user more control. 
Towards this goal, our main idea is to compute a specific type of redundant/over-complete representation of a pre-trained weight matrix and quantize the matrix in that representation. 
We will see how robustness to quantization error will follow 
naturally from our choice of representation.
The de-quantization step uses a straightforward scheme
to re-construct the full-precision weights. We leverage 
a mature concept from Harmonic analysis, Fusion Frames, as the foundation 
for our proposal.

\comment{
{\bf A general roadmap of how to quantize.} Most recent strategies \cite{UrD}
for post-training quantization \cite{OBQ, GPTQ} 
quantize one layer at a time, while minimizing 
a proxy loss. The loss reflects the error incurred due to quantization, 
in that layer's activation (say, on a small calibration dataset). It   
guides the quantization because 
whether a specific weight should be quantized to 
one lattice point or another can impact performance.  
This is based on a clever observation in \cite{UrD} that 
rounding each weight to the nearest quantized lattice 
point is sub-optimal. In any case, the optimization 
task treats the pre-trained weights as a given and 
then seeks to identify the best quantization. Once a specific weight (or a column of weights) has been quantized, the remaining weights may be adjusted 
to compensate, to better match the layer's activation. With this roadmap
in hand, consider an alternative scenario: one where we derived 
a {\em different} representation of the weights. {\color{red}Li: this part is a little bit hard to follow, probably rephrase after the whole paper is done and we settle down on the main pitch.}
Imagine that this representation is such that it is {\em aware} of 
a (potentially sub-optimal) quantization strategy that will follow, which 
will introduce data-dependent noise (e.g., not uniform white noise). 
If this representation is such that it is, {\em by design},  
robust to quantization noise and offers recovery guarantees, then we 
will only minimally compromise on reconstructing the unquantized 
weights faithfully, even if the foregoing quantization roadmap remained 
mostly unchanged otherwise. 
}

{\em Fusion Frames} \cite{720544,OleChristen5009} serve an important role in 
signal processing in analog-to-digital conversion and signal transmission. 
Frames are guaranteed to be robust when the Frame coefficients 
are corrupted by additive noise. They are numerically stable, 
and if additional compute/memory overhead is acceptable, denoising filters with good theoretical properties 
or provably optimal recovery schemes are known. To our knowledge, Frame theory for neural network quantization is unexplored. Our  {\bf key contributions} include
\begin{inparaenum}[\bfseries (a)]
    \item an approach that offers fractional bit quantization capabilities with theoretical guarantees.
    \item We empirically verify that Transformer-based models can be quantized to two bits (or 2.x bits), on an extensive basket of 15 popular Vision Transformers and Large Language Models from the OPT \cite{zhang2022opt} as well as Llama2 \cite{touvron2023llama} classes. We achieve consistent improvements over all existing baselines. 
\end{inparaenum} 

\comment{
{\bf This paper.} 
The starting point of this work is a mature set of 
results from Harmonic analysis, 
specifically related to {\em Frames} \cite{https://ieeexplore.ieee.org/document/720544,https://link.springer.com/book/10.1007/978-3-319-25613-9}, that serve an important role in 
signal processing in analog-to-digital conversion and signal transmission. 
Frames are guaranteed to be robust when the Frame coefficients 
are corrupted by additive noise. They are extremely numerically stable, 
and if some additional compute/memory overhead is acceptable, denoising filters with good theoretical properties 
or provably optimal recovery schemes are known. Our goal 
is to deploy this idea, specifically Fusion Frames, for 
quantizing Vision Transformers. Procedurally, 
we perform the calculations but in the Fusion Frame representation space. 
The de-quantization step uses a straightforward scheme
to re-construct the full-precision weights. 
}

\comment{
Our {\bf key contribution}
is to {\bf (a)} show that Vision transformers can be quantized to two bits (or 2.x bits), with  
only a small drop in performance and that {\bf (b)} the aforementioned behavior 
holds for every single model attempted in our experiments, which includes an 
extensive basket of {\color{red}XX} popular pre-trained models widely 
used in our community. We achieve consistent improvements over all existing baselines. 
}

\subsection{Related Work}


\label{sec:previous work}
Given the growth in the scale 
of foundation models common in our community, model compression is an active topic of research. Distillation 
\cite{hinton2015distilling, zhu2021data}, pruning/shrinking \cite{8485719} and the 
use of sparsity is quite common \cite{6288897, yun2020n}.
There is growing interest \cite{Rokh_2023, namburi-etal-2023-cost, gholami2022survey} in approaches that perform model compression via quantization either (i) during training or (ii) post-training since minimal changes to the architecture are needed. Quantization during training works well  \cite{gholami2022survey, Nagel2021AWP}, but models must be re-trained. 
Post-training quantization (PTQ) methods \cite{nagel2019data} simply quantize a pre-trained model on a small calibration set, and involve much less work. These methods are effective for large language models like OPT \cite{zhang2022opt}, BLOOM \cite{workshop2023bloom} and can reduce the bit-width with only a small degradation in performance. For example, \cite{pmlr-v119-nagel20a} 
analyzed the effect of data-dependent rounding. A layer-wise proxy loss was studied and 
AdaRound quantization was proposed to efficiently minimize this loss. The approach in \cite{frantar2022optimal} minimizes the squared error similar to \cite{pmlr-v119-nagel20a}, 
but quantizes each layer individually while adjusting the remaining unquantized weights using the Hessian of the proxy loss term following \cite{OBDLecunJohn1989, hassibi1993optimal}. 
{OPTQ} \cite{frantar2023optq}(formerly GPTQ) extended upon 
the ideas in OBQ \cite{frantar2022optimal},  
and offered other adjustments that gives a stable scheme that can 
compress large language models like OPT-175B 
or BLOOM-176B to 3 or 4 bits per parameter without a large loss in accuracy.
For Vision Transformers, {PTQ4ViT} \cite{yuan2022ptq4vit} 
quantifies the weights in two stages, 
and uses a Hessian-guided search for the optimal scale for the weights. 
In \cite{liu2021post}, a feature map is used to search for the optimal quantization interval for maintaining similarity between the quantized and original feature maps. The method also chooses different bit widths for each layer.
Other strategies proposed for PTQ include \cite{10.1145/3503161.3547826,li2023repq}.
We note a recent concurrent result for two-bit quantization for language models reported in \cite{chee2023quip}. 
Our approaches are based on different starting points: our choice of 
Frame theory to minimize quantization error versus the choice in \cite{chee2023quip} of using incoherence as a pre and post-processing step, which is later shown to offer desirable theoretical 
properties. But fundamentally, both methods work well due to similar underlying principles 
related to basis expansions (on a space-filling basis). 
We discuss later how \cite{chee2023quip} can be viewed as a special version of 
our formulation (but with no redundancy). 
%
\comment{
  \cite{WF paper from chapter} show that Fusion Frames are robust to noisy coefficients. We use this as a motivation for our approach - modeling the quantization noise as perturbations of the coefficients and leveraging the robustness of Fusion Frames to enable less noisy reconstruction of the signal.
 }
  %



\comment{
is to quantize the model layer by layer while minimizing a proxy loss, which is defined as the norm of error in the activations due to quantization. 
However, the weights are quantized directly in all these methods without any transformations. 
On the other hand,  This is the motivation for our approach. We project the activations onto Fusion Frames and model the quantization noise as being added to the projections. This allows us to utilize the theoretical insights about the robustness of Fusion Frames in the presence of noise. 

\begin{enumerate}[label={(\roman*)}, align=left, left=0pt]
    \item A novel approach to Vision Transformer quantization based on projections onto Fusion Frames
    \item We evaluate out method on image classification task on Imagenet-1k and show that our method performs on-par or better than previous works
\end{enumerate}

The rest of the paper is organized as follows. Section \ref{sec:fusion frames intro} introduces Fusion Frames and describes the robustness of Fusion Frames in the presence of quantization noise. Section \ref{sec:previous work} describes the previous work and how our approach is different from the previously proposed methods. Section \ref{sec:method} describes the proxy loss and our method. \textcolor{blue}{describe other sections as well}.    
}

\section{Finite Frame Theory and Fusion Frames}
\label{sec:fusion frames intro}
Frames generalize the Orthogonal basis decomposition of a Hilbert space and provide redundant representations. Finite frames find applications in robust signal transmission with quantization and erasures \cite{650985, GOYAL2001203, casazza2003equal}, Coding theory \cite{strohmer2003grassmannian}, distributed processing \cite{casazza2008fusion}, Compressed Sensing \cite{ffcsGK2009RH} among others.  
%
We start with a brief review of relevant concepts. Readers familiar with these concepts may skim this section. 

\begin{figure}[tb!]
     \centering
     \includegraphics[width=0.9\linewidth]{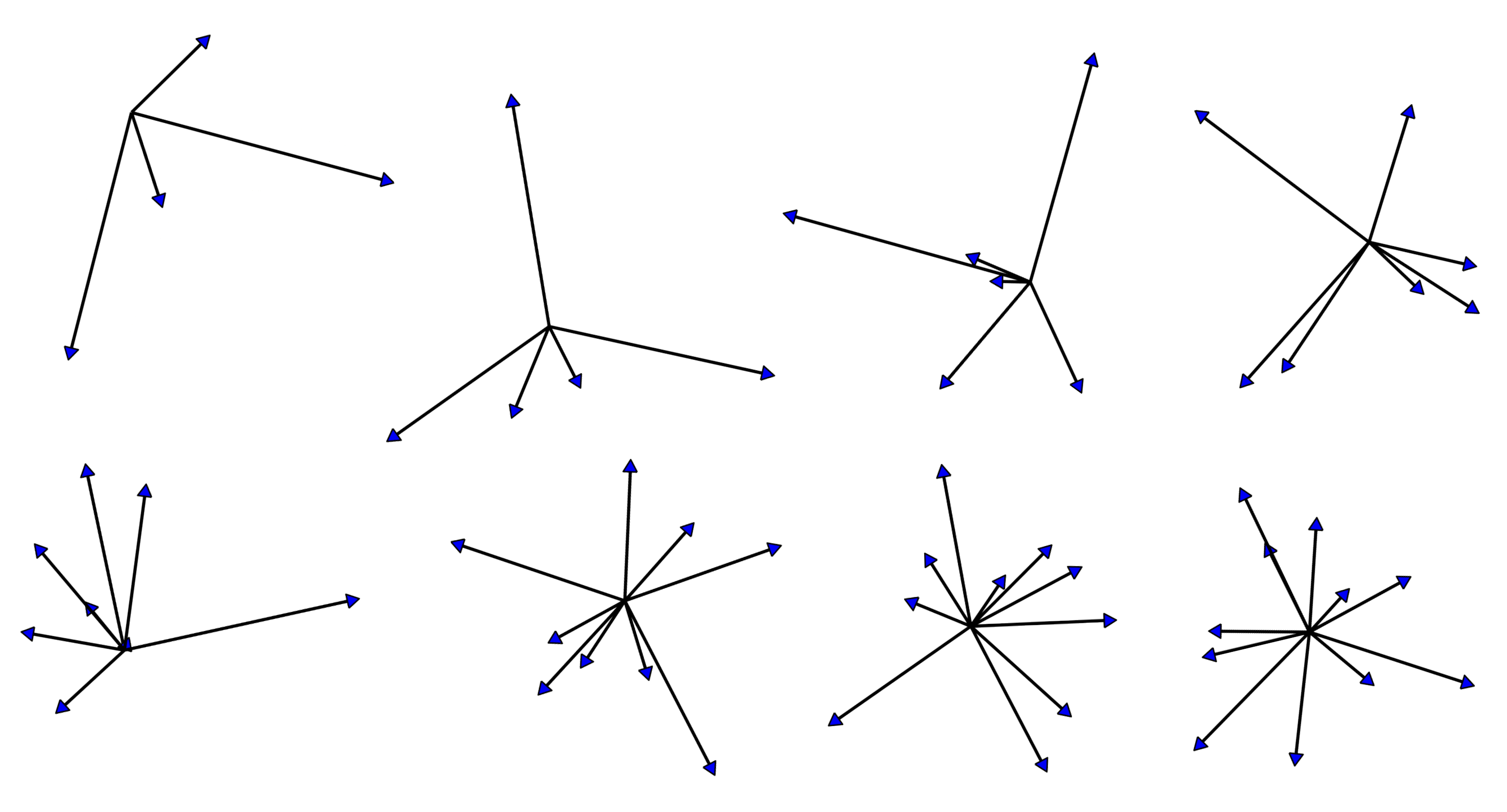}
     \caption{Examples of Tight frames of $k = 4,5,...,11$ in $\mathbb{R}^2$}
     \label{fig:2dframes}
     \vspace{-12pt}
 \end{figure}

Consider a finite-dimensional Hilbert space $\mathcal{H}$ of dimension $d$. Throughout the paper, we denote this space as $\mathcal{H}^d$. 
\begin{definition}[Frames]
 A family of $k$ vectors \textcolor{BlueViolet}{$\upphi = (\varphi_i)_{i=1}^{k}$} in $\mathcal{H}^d$ is called a \emph{frame} for $\mathcal{H}^d$ if there exist constants $0 < A \leq B < \infty$ such that 
\begin{align}
    A||x||^2 \leq \sum_{i=1}^{k} |\langle x, \textcolor{BlueViolet}{\varphi_i} \rangle|^2 \leq B || x ||^2 
\end{align}
for all $x \in \mathcal{H}^d$ where $\langle \cdot, \cdot \rangle$ is the dot-product. The constants $A$ and $B$ are the \emph{lower} and \emph{upper frame bounds}.    
\end{definition}
 The sandwich expression suggests that $x$ will not be poorly distorted when we calculate its inner products with a frame. 
  When $A=B$, \textcolor{BlueViolet}{$\upphi$} is called a \emph{A-tight} frame. When $A=B=1$, we get a Parseval's frame. Fig. \ref{fig:2dframes} shows examples of Tight Frames for $\mathbb{R}^2$ for different $k$'s. The lower bound is equivalent to asking that \textcolor{BlueViolet}{$\upphi$} span $\mathcal{H}$. So, for a frame, we always have $k \geq d$. If $k = 3d$, the redundancy is $r=3$.

Fusion Frames provide a way for fusing ``smaller'' frames to construct large frames, offering various 
efficiency and robustness properties \cite{Eldar2008BeyondBS}.
Formally, 
\begin{definition}[Fusion Frames]
Let $\ffwitok$ be a family of subspaces in $\mathcal{H}^d$, and let $(w_i)_{i=1}^k \subseteq \mathbb{R}^+$   be a family of weights. Then, $\ffwwitok$ is a \emph{fusion frame} for $\mathcal{H}^d$, if there exists constants $0 < A \leq B < \infty$ such that
\begin{align*}
    A||x||^2 \leq \sum_{i=1}^{k} w_i^2|| U_i (x)||^2 \leq B || x ||^2  \quad \text{for all } x \in \mathcal{H}^d
\end{align*}
where $\projectui$ denotes the orthonormal projection onto the subspace $\ffwi$ for each $i$. The constants $A$ and $B$ still denote the \emph{lower} and \emph{upper fusion frame bounds} respectively. 
\end{definition}
Similar to the Frames case, the Fusion Frame $\ffwwitok$ is referred to as a \emph{tight fusion frame} if $A=B$ and as a \emph{Parseval fusion frame} if $A=B=1$. Finally, if $w_i = 1$ for all $i$, we simply utilize the notation $\ffwitok$.

\subsection{Operators in Fusion Frames}
\label{sec:Operators in Fusion Frames}

Fusion Frame (FF) operators can be formally defined using a \emph{Hilbert direct sum}. Since we use the operators for model quantization, without loss of generality, we describe them in terms of vectors and matrices, to keep notations simple. Let $\ffwwitok$ be a Fusion Frame for $\mathcal{H}^d$ with {\color{BlueViolet}orthonormal basis}  $\orthopitok$ respectively.

The \emph{Analysis operator} $\mathcal{T}_{\mathcal{W}}$ 
takes a signal $x \in \mathcal{H}^d$ and {\color{Green}computes its dot product with all the basis $\orthopitok$.} \comment{provides 
a sequence of {\color{Green}vectors} in $\ffwitok$.} 
The results represent 
$x$ w.r.t. the FF as 
\begin{equation}
\mathcal{T}_{\mathcal{W}} :  x \mapsto (w_i \textcolor{Green}{P_i^T (x)})_{i=1}^k
\end{equation}
The \emph{Synthesis operator} $\mathcal{T}^*_{\mathcal{W}}$ is the adjoint of the analysis operator, 
and takes a sequence of representation vectors $\textcolor{Green}{{(y_i)}_{i=1}^k}$ 
and outputs a signal in $\mathcal{H}^d$: the reconstruction of the original signal from its {\color{Green}FF representation} is defined as
\begin{equation}
\mathcal{T}^*_{\mathcal{W}} : \textcolor{Green}{(y_i)}_{i=1}^k \mapsto \sum_{i=1}^k w_i \orthopi (\textcolor{Green}{y_i})
\end{equation}
{The \emph{Fusion Frame operator} $\mathcal{S}_{\mathcal{W}}$ is defined as the {\em composition} of these two operators. It first computes the FF {\color{Green}representation} of a signal in $\mathcal{H}^d$ in {\em different} {\color{blue}subspaces} using the Analysis operator. Then, when needed, we can reconstruct the signal back from these representations using the Synthesis operator. When the Fusion Frame is tight, the reconstruction is exact \cite{CASAZZA2011175}. Formally,}
\begin{equation}
\mathcal{S}_{\mathcal{W}} = \mathcal{T}^*_{\mathcal{W}}\mathcal{T}_{\mathcal{W}}: x \mapsto \sum_{i=1}^k w_i^2 \textcolor{cyan}{U_i (x)}
\end{equation}
Here, $\projectui = \orthopi \orthopi^T$ is the orthogonal projection onto the subspace $\ffwi$. If the Fusion Frame is tight, we have $\mathcal{S}_{\mathcal{W}} = AI_d$ where $I_d$ is the $d \times d $ Identity Matrix. Throughout, we will use Parseval Fusion Frames, where the frame bounds $A=B=1$. 
Fusion Frames offer many other properties but due to space, we will
keep the presentation focused.

\noindent {\bf How will Fusion Frames be used?} 
An easy way to see Fusion Frames in practice is to work out a simple example, 
\begin{example}
\label{ex:ff}
Consider the Euclidean space $\mathcal{H}^d = \mathbb{R}^4$. Say, an oracle gives us a Fusion Frame where we have $k = 3$ subspaces, and each subspace is of equal dimension $\rho=2$. For notational ease, we represent these subspaces with their Synthesis operator 
$\mathcal{T}^*_{\mathcal{W}}$
{\small\begin{equation*}    
 = \left( 
 \textcolor{BlueViolet}{
\begin{bmatrix}
 0.57 &  0.00 \\
 0.00 &  0.57 \\
 0.57 &  0.00 \\
 0.00 &  0.57
\end{bmatrix}, }
\textcolor{BlueViolet}{
\begin{bmatrix}
 0.57 &  0.00 \\
0.00 &  0.57 \\
-0.28 &  0.50 \\
-0.50 & -0.28
\end{bmatrix}, }
\textcolor{BlueViolet}{
\begin{bmatrix}
 0.57 &  0.00 \\
 0.00 &  0.57 \\
-0.28 & -0.50 \\
 0.50 & -0.28
\end{bmatrix}}
\right)
\end{equation*}
}
We want to compute the FF representation of a signal $x = 
\begin{bmatrix}
-1 & -0.5 & 0.5 & 1
\end{bmatrix}^T
$. 
To do so, we must apply the Analysis operator $\mathcal{T}_{\mathcal{W}}$ on $x$.
The Analysis operator is simply based on the 
individual transposes in  $\mathcal{T}^*_{\mathcal{W}}$ defined above. 
{\small\begin{equation*}    
\textcolor{BlueViolet}{
\begin{bmatrix}
0.57 & 0.00 & 0.57 & 0.00 \\
0.00 & 0.57 & 0.00 & 0.57
\end{bmatrix}, }
\textcolor{BlueViolet}{
\begin{bmatrix}
0.57 & 0.00 & -0.28 & -0.50 \\
0.00 & 0.57 & 0.50 & -0.28
\end{bmatrix}}
\cdots
\end{equation*}
}

{Applying $\mathcal{T}_{\mathcal{W}}$ on $x$, we get the FF representations}
\begin{equation*}
 \textcolor{Green}{\mathcal{T}_{\mathcal{W}} (x)} = \left( 
 \textcolor{Green}{
\begin{bmatrix}
-0.28\\ 
 0.28
\end{bmatrix}, }
\textcolor{Green}{
\begin{bmatrix}
-1.22\\
-0.32
\end{bmatrix}, }
\textcolor{Green}{
\begin{bmatrix}
-0.22\\
-0.82
\end{bmatrix}}
\right)  
\end{equation*}
To get the actual projections of $x$ onto different subspaces $\ffwi$, we multiply these coefficients with the scaled orthonormal basis $(w_i \orthopi)_{i=1}^k$ of their corresponding subspaces 
\begin{equation*}
    (w_i^2 \textcolor{cyan}{U_i (x)})_{i=1}^3 = \left( 
    \textcolor{cyan}{
    \begin{bmatrix}
        -0.1667 \\
         0.1667 \\
        -0.1667 \\
         0.1667
    \end{bmatrix}}, 
    \textcolor{cyan}{
    \begin{bmatrix}
        -0.7053 \\
        -0.1890 \\
         0.1890 \\
         0.7053
    \end{bmatrix} },  
    \textcolor{cyan}{
    \begin{bmatrix}
        -0.1280 \\
        -0.4777 \\
         0.4777 \\
         0.1280
    \end{bmatrix} } 
    \right)
\end{equation*}
We can verify by checking the identity 
$\mathcal{S}_{\mathcal{W}} = I_d$ or checking that $\sum_{i=1}^3 w_i^2\textcolor{cyan}{U_i (x)} = x$ (only accurate up to rounding errors) that this Fusion Frame is a Parseval's frame.
Applying the Synthesis operator $\mathcal{T}^*_{\mathcal{W}}$ on the projections above recovers $x$ perfectly.
\label{ex: tff reps}

\textbf{Corrupting FF representations by noise.} 
 What happens when the Fusion frame representations are corrupted by noise, say due to erasure or quantization? Because of redundancy in the 
 {\color{Green}representation of a signal}, we expect some immunity to corruptions in the representations due to noise. 
 In the current example, this is indeed the case. 
 If we add noise to \textcolor{Green}{$\mathcal{T}_{\mathcal{W}}(x)$} with an SNR of $10$\text{dB} and use the noisy coefficients to reconstruct $x$ back, we observe an MSE reduction of $33\%$ at a redundancy factor of $r=1.5\times$ and $50\%$ MSE reduction $r=2\times$, consistent with theory \cite{650985}. 
\end{example}

{\bf Quantizing Transformer layers.} Let us consider quantizing each layer in a 
Transformer model as in \cite{pmlr-v119-nagel20a, frantar2022optimal, frantar2023optq, yuan2022ptq4vit}, e.g., by quantizing individual weights or columns, one by one. First, notice that the quantization error/noise is weight-dependent. Further, the error will also 
depend on how all other weights are quantized. The only 
way to guide a quantization scheme 
is the evaluation of a loss (to be described shortly) on a small calibration dataset $\mathcal{D}$. In this regime, even with strong assumptions on the noise, it is difficult to say much about the quality 
of the de-quantization. 
On the other hand, far more is known \cite{650985, shWDFFTF, FFTaAPGCGKB2012} about the behavior of quantization of data given in an appropriate Frame basis (e.g., Fusion Frames), 
and error bounds on the reconstruction are available. 
Put simply, quantization noise in the space of Frame projections incurs far less error in the reconstructions due to the robustness of Frame 
representations. \S \ref{sec:method} will leverage this principle.


\subsection{Tight Fusion Frames and their construction}
\label{sec:tff_construction}
\comment{Since we focus on Transformer-based models, we know that the parameters and the activations of the model are real-valued. So, it is safe to 
restrict to Hilbert spaces of the form $\mathbb{R}^d$. In Example \ref{ex:ff}, a Fusion Frame was assumed to be provided.
However, off-the-shelf deterministic 
algorithms \cite{CASAZZA2011175} are readily available. }We first define the type of Fusion Frames we will use and then describe how they can be constructed. 

\begin{definition}[Tight Fusion Frames or TFF]
    For 
    $A>0$ and with $I_d$ giving the $d\times d$ Identity matrix, 
    a $(k,\rho,d)$-TFF is a sequence $\{\projectui \}_{i=1}^k$ of $d \times d$ orthogonal projection matrices of rank $\rho$ and scalars $\{w_i\}_{i=1}^k, w_i > 0$ such that 
    \vspace{-7pt}
    \begin{equation}
    \sum_{i=1}^k w_i^2 \projectui = AI_d.
    \vspace{-10pt}
    \end{equation}
\end{definition}
A $(k,\rho,d)$-TFF is a sequence of $k$ equidimensional sub-spaces of dimension $\rho$ in a $d$-dimensional space, and $\projectui$ is the orthogonal projection matrix onto the $i^{th}$ sub-space.

{\bf Constructing TFFs.}
The algorithm in \cite{CASAZZA2011175} can be used to 
generate TFFs if we provide 
the dimension $d$, the number $k$ of subspaces we need, and the dimension $\rho$ of each of these subspaces.\comment{The low-level details of the 
algorithm are not important to our discussion
since it can be used as a black-box module (a brief synopsis is still included in \S\ref{sec:construct tff}). 
Nonetheless, we give a very brief summary for 
the interested reader.} The algorithm has 
two main steps. First, one generates a Tight Frame of $d$ unit norm vectors for the complex domain $\mathbb{C}^\rho$. Then, this Frame is modulated with the square roots of unity to generate the $k$ subspaces for $\mathbb{C}^d$.  
We use a simple construction 
described in 
\cite{FICKUS20231}
to extend these Fusion Frames to $\mathbb{R}^d$. Since it can be used as a black-box module, we skip the details and include a brief synopsis in Appendix \S\ref{sec:construct tff}.

{\em Remarks.} 
A few properties are useful to note. 
This Fusion Frame construction is sparse/block diagonal and can be generated one subspace at a time. To generate another Fusion Frame, we can hit it with a random rotation. Depending on 
the Transformer model at hand, the dimension of the activations of the layer determines $d$. For a desired redundancy factor ($k \times \rho \geq d$) in our frames, given $d$ we simply choose a $k$ and $\rho$ such that they are valid (i.e., a TFF exists for the triple $(k,\rho,d)$) according to \cite{CASAZZA2011175}. If not, we use a slightly lower redundancy factor $r$ knowing that we 
will always have a trivial solution for $k=1$ and $\rho = d$.


\section{Fusion Frames based Quantization}
\label{sec:method}
We can now leverage the ideas described 
in the preceding sections for quantizing the parameters of a Transformer model. Consistent with common PTQ approaches \cite{pmlr-v119-nagel20a, frantar2022optimal, frantar2023optq, yuan2022ptq4vit}, we perform quantization layer-by-layer, minimizing the proxy loss between the quantized and non-quantized output of the layer.

\begin{figure}[!tb]
    \centering
    \includegraphics[width=0.995\linewidth]{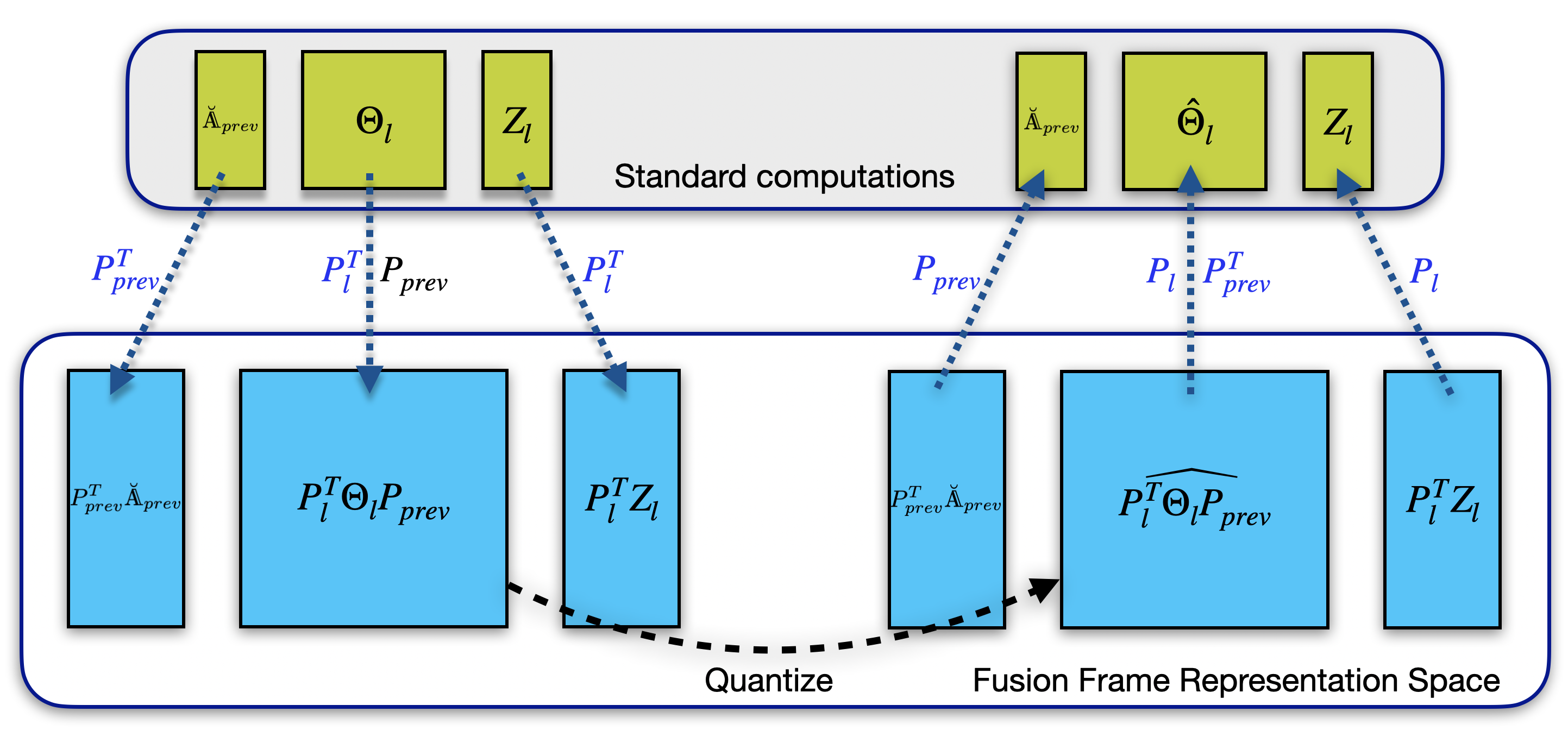}
    \vspace{-15pt}
    \caption{{Illustration of standard calculation (on top) versus the corresponding calculations in FF space (bottom)}}
    \label{fig:enter-label}
    \vspace{-15pt}
\end{figure}
{\bf What are analogous calculations in FF space?} Consider a layer $l$ in a Transformer model, with parameters $\Theta_l$. Let ${\mact{A}}_{\rm prev}$ be the activation of the already quantized previous layer for the examples in the calibration set $\mcal{D}$. 
The (non-quantized) output $Z_l$ of layer $l$ is 
\begin{equation}
\label{eq:ZisthetaA}
    {Z}_l = \Theta_l {\mact{A}}_{\rm prev}
\end{equation}
Here, $\Theta_l$ maps the input ${\mact{A}}_{\rm prev}$ to the output $Z_l$. To avoid directly quantizing $\Theta_l$, we want the quantization noise to instead impact the analogous terms in the Fusion Frame representation (but equivalent calculation as \eqref{eq:ZisthetaA}). To this end, let us set up some notations. In general, the dimension of $Z_l$ and ${\mact{A}}_{\rm prev}$
may not be the same. So, the number of subspaces in their 
respective Fusion Frames will 
be different. Let $k, k_{\rm prev}$ denote the number of subspaces for $Z_l$ 
and ${\mact{A}}_{\rm prev}$ respectively. In other words, 
$\ffwl = \ffwilitok $ and 
$\ffwprev = \ffwiprevitok $.  Let the sequence of orthonormal basis for the subspaces of $\ffwl$ and $\ffwprev$ be given by $\textcolor{BlueViolet}{(P_i^{l})}_{i=1}^{k}$ and $\textcolor{BlueViolet}{(P_i^{\rm prev})}_{i=1}^{k_{\rm prev}}$ respectively. To reduce notational clutter, we absorb the scalars $w_i$ into $\textcolor{BlueViolet}{P_i}$. 
%
To write down the expression in FF space,
for simplicity, let us vectorize the set of 
orthogonal basis above and define
\begin{equation*}
\textcolor{BlueViolet}{P_{l}} = [\textcolor{BlueViolet}{P^{l}_1} \textcolor{BlueViolet}{P^{l}_2 \dots P^{l}_{k}}] \; \text{and} \; \textcolor{BlueViolet}{P_{\rm prev}} = [\textcolor{BlueViolet}{P^{\rm prev}_1 P^{\rm prev}_2 \dots P^{\rm prev}_{k_{\rm prev}}}] 
\end{equation*}

Taking the FF representations of the output $Z_l$ means
\begin{equation}
    \textcolor{Green}{P_l^T Z_l} = \textcolor{BlueViolet}{P_l^T} \underbrace{\Theta_l {\mact{A}}_{\rm prev}}_{=Z_l} 
\end{equation}
Rearranging brackets, 
\begin{align}
    \textcolor{BlueViolet}{P_l^T} \Theta_l {\mact{A}}_{\rm prev} 
    &= \textcolor{BlueViolet}{P_l^T}  \Theta_l (\textcolor{BlueViolet}{P_{\rm prev}  P_{\rm prev}^T} ) {\mact{A}}_{\rm prev} \\
    &= (\textcolor{BlueViolet}{P_l^T}  \Theta_l \textcolor{BlueViolet}{P_{\rm prev}} ) ( \textcolor{Green}{P_{\rm prev}^T {\mact{A}}_{\rm prev}} )
\end{align}
In the above expression, 
the object $(\textcolor{BlueViolet}{P_l^T}  \Theta_l \textcolor{BlueViolet}{P_{\rm prev}} )$  maps the FF representation of $\mact{A}_{\rm prev}$, i.e., $(\textcolor{Green}{P_{\rm prev}^T {\mact{A}}_{\rm prev}} )$, 
to the FF representation of $(\textcolor{Green}{P_l^T Z_l} )$. 
This operation is completely in the FF representation space as desired.

\comment{
{\color{cyan} L: for a redundancy factor of 1.1, is it equivalent that we use 2.2 bits (instead of 2 bits) for quantization, from a model size pov? H: Yes. Other methods can only jump like 2 bits to 3 or 4 bits, but we can do something in between as well. L: great we probably should high light this property. H: Thats a good point, we will mention it in the paper. L: also. in low bit, this fractional capability is particularly important. H: Yeah, makes sense. L: if we want to call out 2.x capability, it might be useful to have a few different x values to shown the quantization quality changes quite a bit. is it possible to run the experiments for supplemental material? V: Li, yes actually ablation with redundancy will go 
into the main paper itself.}
}
A notation simplification allows us to cross-reference what our FF-space calculations are doing w.r.t. the objective function. Let $\textcolor{Green}{C_{\rm prev}}  = \textcolor{Green}{P_{\rm prev}^T {\mact{A}}_{prev}} $ and $D_l = \textcolor{BlueViolet}{P_{l}^T}  \Theta_l \textcolor{BlueViolet}{P_{\rm prev}} $. Our objective is to quantize $D_l$ to $\hat{D}_l$ while minimizing the proxy loss in terms of FF representations,
\begin{align*}
    \mathcal{L}(\hat{D}_{l}) &= ||D_l \textcolor{Green}{C_{\rm prev}} - \hat{D}_{l} \textcolor{Green}{C_{\rm prev}}  ||_F^2 \\
    &= \text{tr}((D_l - \hat{D}_l)^T \textcolor{Green}{C_{\rm prev}^T C_{\rm prev}} (D_l - \hat{D}_l)) \\
    &= \text{tr}((D_l - \hat{D}_l) \textcolor{Green}{C_{\rm prev} C_{\rm prev}^T} (D_l - \hat{D}_l)^T )
\end{align*}
The term $\tilde{H} = \textcolor{Green}{C_{\rm prev} C_{\rm prev}^T} $ corresponds to the Hessian prominent in most published results on PTQ strategies \cite{pmlr-v119-nagel20a, frantar2022optimal, frantar2023optq, chee2023quip}. 
So, our loss is the same as other approaches, except that we are operating in the FF representation space and enjoy all the associated noise robustness properties. Further, because the loss for quantizing the transformed weights $D_l$ is the same as e.g., \cite{frantar2023optq}, we can directly use the Hessian-based iterative quantization algorithms in \cite{frantar2022optimal, frantar2023optq} with minimal changes. 
Finally, following recent results in Post-training Quantization \cite{pmlr-v119-nagel20a, frantar2022optimal, frantar2023optq, chee2023quip}
we primarily focus on quantizing the transformed weights $(D_l)$ but 
include one experiment with a simple activation quantization in \S\ref{sec:Activation Quantization}. We note that there are standalone activation quantization strategies for smaller Vision models for up to four bits, see  \cite{10.1145/3503161.3547826, yuan2022ptq4vit}.
\begin{figure}[!bt]
    \centering
    \includegraphics[width=0.7\linewidth]{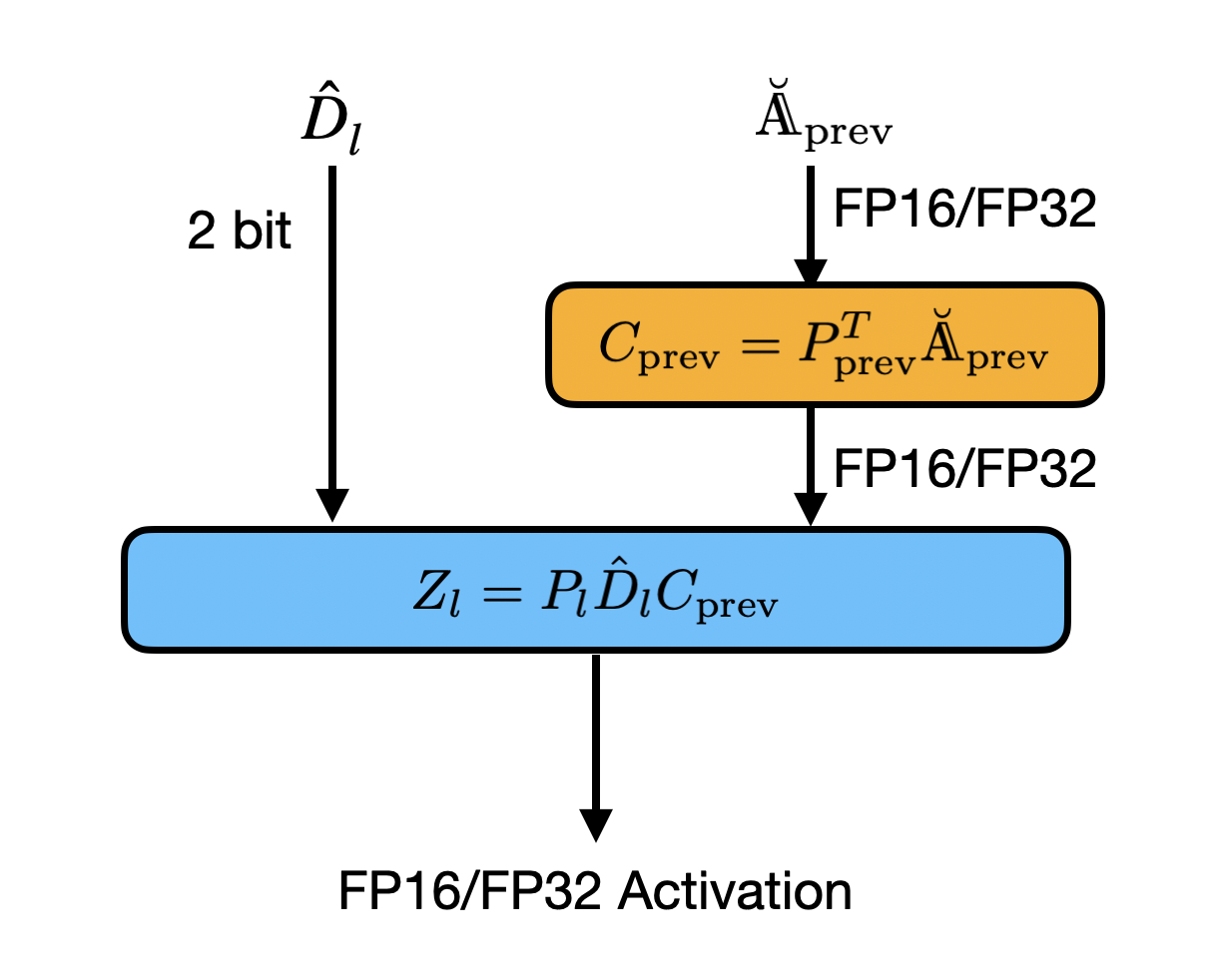}
    \vspace{-10pt}
    \caption{Inference for a FrameQuant quantized model.}
    \label{fig:inference}
    \vspace{-11pt}
\end{figure}

\textbf{Details of the quantization procedure.}
Other than working in the FF space, the quantization itself is 
almost identical to \cite{frantar2023optq}. We use the iterative method from \cite{frantar2023optq} with some modifications to improve the stability of our algorithm. For example, we found that clipping the weights before calling the iterative scheme from 
GPTQ reduces the weight range during quantization. This effectively adds more quantization noise to the outlier weights that are too large. Since Fusion Frames spread out the energy uniformly among different subspaces, we observe that there are only a few outliers in the transformed Weight matrices, and hence clipping them boosts performance. 
We found that simply clipping the weights at $2\sigma$ (assuming a Normal distribution), where $\sigma$ is the standard deviation of $D_l$, works well in practice. We observe that this change also helps the method in \cite{chee2023quip} (and this modified algorithm is also included in our baselines). Alg. \ref{alg:FFQuant} shows the sequence of steps in FrameQuant. 
\begin{algorithm}
\begin{algorithmic}
\REQUIRE Weight matrix $\Theta_l$, previous layer activations ${\mact{A}}_{prev}$, input and output Fusion Frames $\textcolor{BlueViolet}{P_l}, \textcolor{BlueViolet}{P_{\rm prev}}$, block size $B$
\STATE 1: Compute $\textcolor{Green}{C_{\rm prev}} = \textcolor{Green}{P_{\rm prev}^T A_{\rm prev}}$, $D_l = \textcolor{BlueViolet}{P_{l}^T} \Theta_l \textcolor{BlueViolet}{P_{\rm prev}}$
\STATE 2: Compute $\sigma = std(D_l), \;\;\; \mu = mean(D_l)$
\STATE 3: $D_l = 2\sigma \text{ clip}(D_l, \mu - 2\sigma, \mu + 2\sigma)$
\STATE 4: $\hat{D}_l = quantize(D_l, \textcolor{Green}{C_{\rm prev}}, B)$ \quad  // modified GPTQ
\STATE 5: Store $\hat{D}_l$ \quad \quad \quad \quad \quad // store the quantized matrix $\hat{D}_l$
\STATE \textbf{return} $\textcolor{BlueViolet}{P_l} \hat{D}_l \textcolor{Green}{C_{\rm prev}}$ \quad // return quantized layer activations
\end{algorithmic}
\caption{FrameQuant}
\label{alg:FFQuant}
\end{algorithm}
\vspace{-11pt}

\subsection{Robustness of Fusion Frames}
\label{sec: Robustness of Fusion Frames}
We now state some technical results that 
apply to both Frames and Fusion Frames. 
\begin{asparaenum}[\bfseries (a)]
\item {\em Redundancy related guarantees.} During quantization, the Fusion Frame coefficients are corrupted. This can be modeled as an additive noise being added to these coefficients. Assume that 
the redundancy factor is $r>1$. Even 
with classical analysis, 
the result in \cite{anrreudprcj2005, 650985} shows that when using Tight Frames to reconstruct the signal from noisy coefficients, for memoryless quantization, 
we get an MSE reduction of $\mathcal{O}(1/r)$. A rate of $\mathcal{O}(1/r^2)$ for 
{\em consistent} reconstruction can also be achieved
by solving an LP during the dequantization step \cite{650985}. While this may not be preferred in practice, we know that if adopted, this matches the 
lower bound of $(1/r^2)$, see Ch. 2 in \cite{650985}.
\item Another benefit of Frame representations 
is that reconstruction can ``denoise'' using filters available 
in closed form. For example, with Tight Frames, 
it is known that the Wiener filter provably minimizes the MSE, see Ch. 13 in \cite{FFTaAPGCGKB2012}, \cite{KUTYNIOK200964}. In our experiments, we found that even a diagonal approximation of the Wiener 
filter helps. But our experimental results are reported 
without utilizing this boost. 
\end{asparaenum}

\subsection{Inference Procedure}
During inference, the quantized model is loaded into memory. At each layer, the inputs to the layer (${\mact{A}}_{prev}$) are first transformed into their Fusion Frame representations using the analysis operator $\textcolor{BlueViolet}{P^T_{\rm prev}}$. The FF representations are then transformed by the quantized weights ($D_l$) for this layer into the FF representations of the output. Finally the synthesis operator $\textcolor{BlueViolet}{P_{l}}$ is used to compute the layer outputs. 
Figure \ref{fig:inference} shows this dequantization process and the bit widths of each of these operations for a single layer in a network.

\begin{table*}[!ht]
\centering
\setlength{\tabcolsep}{5pt}
{\footnotesize
\begin{tabular}{cccccccccccc}
\toprule
\textbf{Method} & \textbf{\#bits} & \multicolumn{4}{c}{\textbf{ViT}} & \multicolumn{3}{c}{\textbf{DeiT}} & \multicolumn{3}{c}{\textbf{Swin}} \\
& & \textbf{T}& \textbf{S}& \textbf{S/32}& \textbf{B}& \textbf{T}& \textbf{S}& \textbf{B}& \textbf{S}& \textbf{B}& \textbf{B/384} \\
\midrule
Full-Precision & 32 & 75.46 & 81.39 & 75.99 & 85.10 & 72.16 & 79.85 & 81.98 & 82.88 & 84.67 & 86.02 \\
\midrule
PTQ4ViT & 2 & 0.33 & 0.55 & 0.71 & 0.41 & 1.51 & 4.47 & 25.54 & 12.54 & 0.15 & 0.15 \\
GPTQ    & 2 & 0.40 & 0.40 & 0.39	 & 29.26 & 1.60 & 4.23 & 41.00	 & 43.54	 & 47.38	 & 57.52 \\
QuIP    & 2 & 1.42	& 21.98	& 19.00	& 77.54	& 12.93	& 51.62	& 75.51	& 71.58	& 74.91	& 79.85 \\
QuIP (with our $2\sigma$ clip)   & 2 & 9.10	& 48.96	& 41.41	& 79.54	& 30.49	& 65.70	& 77.69	& 76.34	& 79.17	& 82.40 \\
\rowcolor{Gray}
FrameQuant ($r = 1.0$) & 2 & 8.92	& 48.10	& 41.16	& 79.53	& 31.73	& 66.35	& 77.62	& 77.91	& 80.16	& 82.44 \\
\rowcolor{Gray}
FrameQuant ($r = 1.1$) & 2.2 & \textbf{25.79}	& \textbf{61.51}	& \textbf{53.85}	& \textbf{80.93}	& \textbf{46.48}	& \textbf{70.43}	& \textbf{78.67}	& \textbf{78.77}	& \textbf{81.33}	& \textbf{83.42} \\
\midrule
PTQ4ViT & 3 & 18.32 & 36.18 & 22.20 & 21.43 & 51.73 & 69.65 & 75.35 & 73.01 & 69.46 & 70.68 \\
\bottomrule
\end{tabular}
}
\vspace{-0.05in}
\caption{ImageNet-1k Top-1 validation accuracy of Tiny to Base sized Vision Transformer-based models when quantized to 2 (or 3) bits by different methods. FrameQuant with a redundancy factor of $r = 1$ already performs better or on par with Quip \cite{chee2023quip}. With a slightly higher redundancy factor of $r = 1.1$, we get the best performance of all the methods under consideration.}
\label{tab:imagenet1k_val_acc}
\end{table*}

\section{Experiments}
\label{sec: Experiments}

We performed an extensive set of 
experiments comparing FrameQuant with several quantization baselines 
for Vision models and Language models.  
The goal is to assess \begin{inparaenum}[\bfseries (a)]
\item performance metrics of different methods on benchmark tasks and  
\item how close low-bit quantization can approach the full precision performance with a small degree of representation redundancy. 
\end{inparaenum}
We use image classification task \cite{deng2009imagenet} for Vision models and Perplexity for Language models. 

We start with an overview of our experimental setup. We present the evaluation results of FrameQuant on 15+ Vision Transformer architectures+configurations for image classification. Next, we conduct an ablation study on image classification task to better understand the behavior of different components of FrameQuant. We then present results on Language models such as OPT \cite{zhang2022opt} and Llama2 \cite{touvron2023llama} by comparing perplexity and accuracy in downstream tasks. 
The appendix includes many additional experiments. 

\subsection{Experimental Setup}

We evaluate our method on the ImageNet-1K classification task. For quantizing the model weights of the pre-trained models obtained 
from the Huggingface hub \cite{rw2019timm}, we use 
128 images randomly selected images from the training dataset as calibration dataset $\mathcal{D}$. 
We quantize the parameter matrices of the layers sequentially from shallow layers to deep layers, similar to \cite{frantar2023optq}. After quantizing each layer, we pass the inputs to the layer again and send the output with the quantized weights to the next layer. Finally, we evaluate the quantized models on the ImageNet-1K validation dataset and report the top-1 accuracy. All our ``base'' 
experiments correspond to $2$ bits. 
We note that one of the baselines, PTQ4ViT \cite{yuan2022ptq4vit}, performs activation quantization 
together with weight quantization, but was not tested in 
the extreme $2$ bit quantization setting. To ensure fair comparisons to that method, we switch off activation quantization in their method and 
also add another experiment with $3$ bits. For additional experiments with activation quantization, Segmentation and Object Detection tasks, we refer the reader to the Appendix sections \ref{sec:Activation Quantization}, \ref{sec: seg and obj det} respectively.

\begin{table}[tb!]
\centering
\setlength{\tabcolsep}{1.75pt}
{\footnotesize
\begin{tabular}{ccccccc}
\toprule
\textbf{Method} & \textbf{\#bits} & \multicolumn{2}{c}{\textbf{ViT}} & \multicolumn{2}{c}{\textbf{DeiT III}} & \multicolumn{1}{c}{\textbf{Swin}} \\
& & \textbf{L}& \textbf{H}& \textbf{L}& \textbf{H}& \textbf{L} \\
\midrule
Full-Precision &32 &  85.84 & 87.59 & 86.97 & 87.19 & 85.95 \\
\midrule
PTQ4ViT & 2 & 37.05 & 00.18 & 2.14 & 55.57 & - \\
GPTQ    & 2 & 63.08	& 42.63 & 68.43	& 28.20 & 71.69 \\
QuIP & 2 & 82.22 & 	84.58 & 84.76 & 	86.27 & 83.61 \\
QuIP (our $2\sigma$ clip) & 2 & 83.17 & 	85.31 & 85.48 & 	86.38 & 84.27 \\
\rowcolor{Gray}
FrameQuant ($r=1.0$)  & 2 & 83.22 & 	85.49 & 85.45 & 	86.62 & 84.25 \\
\rowcolor{Gray}
FrameQuant ($r=1.1$) & 2.2 & \textbf{83.67} & 	\textbf{85.99} & \textbf{85.75} & 	\textbf{86.68} & \textbf{84.42} \\
\midrule
PTQ4ViT & 3 & 81.26 & 78.92 & 83.63 & 85.39 & - \\
\bottomrule
\end{tabular}
}
\caption{ImageNet-1K Top-1 validation accuracy of Large and Huge sized Vision Transformer-based models when quantized to 2 (or 3) bits by different methods.}
\vspace{-20pt}
\label{tab:imagenet1k_val_acc_large_huge}
\end{table}

\subsection{Results on ImageNet Classification Task} 
\begin{figure*}
    \centering
    \vspace{-0.05in}
    \begin{subfigure}{0.75\linewidth}
        \centering
        \includegraphics[width=\linewidth]{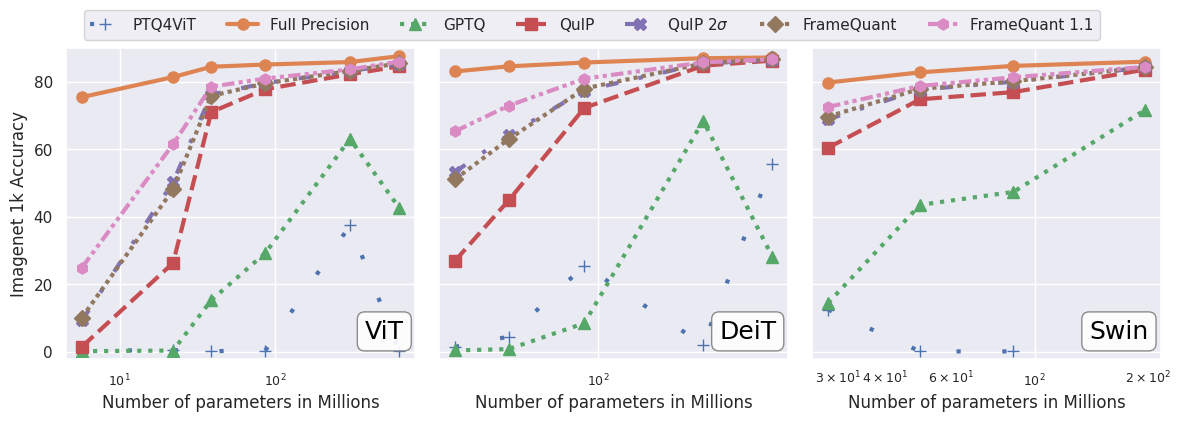}
        \caption{Validation accuracies for different classes of Transformer models for Vision on ImageNet-1K}
        \label{fig:Validation accuracies on Imagenet}
    \end{subfigure}%
    \begin{subfigure}{0.25\linewidth}
        \centering
        \includegraphics[width=\linewidth]{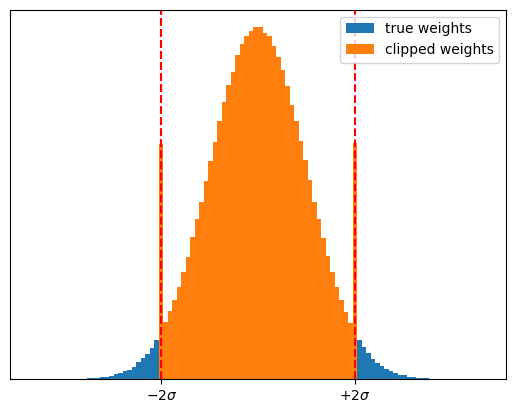}
        \vspace{-6pt}
        \caption{Weights distribution in a ViT layer and the $2\sigma$ thresholds}
        \label{fig:two-sigma}
    \end{subfigure}%
    \vspace{-0.1in}
    \caption{\textbf{(a)} Validation accuracies of Vision Transformers on ImageNet-1K dataset. We can see FrameQuant closing the gap between the full precision model with increasing redundancy. Each dot in the plot corresponds to a model from tables 1-2 combined. \textbf{(b)} shows the distribution of weights in a ViT layer and the $2\sigma$ thresholds for clipping. We see that our thresholding keeps most of the mass while removing outliers. }
    \label{fig:ViT Val acc}
    \vspace{-0.1in}
\end{figure*}

We use model architectures (including ViT \cite{dosovitskiy2021an}, DeiT \cite{pmlr-v139-touvron21a}, DeiT III \cite{touvron2022deit}, and Swin \cite{liu2021swin}) and model sizes (including small, medium, large, huge) that are available on the Huggingface hub \cite{rw2019timm}. Our main 
results for these experiments are shown in Tab. \ref{tab:imagenet1k_val_acc}--\ref{tab:imagenet1k_val_acc_large_huge}. Figure \ref{fig:Validation accuracies on Imagenet} shows the performance of the different classes of models on the ImageNet-1K dataset.
 We observed that clipping the weights at $2\sigma$ also helps QuIP \cite{chee2023quip}, so we include it as an additional baseline.
 Even with a redundancy factor of $r=1$, FrameQuant achieves better accuracy compared to most baselines under consideration. 
Further, with a redundancy factor of $r = 1.1$, FrameQuant outperforms all baselines by a good margin and is respectably close to the full precision model, underscoring the robustness of Fusion Frames in the presence of quantization noise. We observe that adding more redundancy to the Frame representations continues to improve the performance of the quantized models, especially when the models are small. See \S\ref{sec:Impact of redundancy in representations} for more details. We note that the codebase for PTQ4ViT \cite{yuan2022ptq4vit} 
was not compatible with the Swin-L model, so we could not report their performance for this model.


\subsection{Ablation Study}
\label{sec: ablation study}
In this section, we dissect FrameQuant to understand the contribution of different components of our algorithm. Table \ref{tab: ablation study} shows the results of this experiment. We use GPTQ \cite{frantar2023optq} as our starting point. With GPTQ \cite{frantar2023optq} alone, the performance drops in the quantized models are significant: as high as $82\%$ for the DeiT III \cite{touvron2022deit} Base model. Simply with the FF representation added (column \textbf{TFF}), we see improvements in performance across all models, with a maximum improvement of $56\%$ for DeiT III-H. We note that some of the smaller-sized models are yet to see all the benefits of FF representations. That is because these models have 
outliers in the weights (much larger than the remaining weights) which results in higher quantization errors. 
The FF representation yields a nice enough distribution 
that we can fit a Normal distribution. 
So, after we clip these weights at the $\pm 2\sigma$ level, we see improvements in performance because of the outlier removal. Clipping is most effective once the weights are nicely distributed. A direct application of clipping on the weights has limited efficacy and incurs errors for weight configurations that are degenerate/poorly distributed, see \ref{sec:gptq+clip} for more details.
Finally, we add a redundancy factor of $r=1.1$ and the FF representations take advantage of this redundancy: we see the best performance across the board.

\begin{table*}[ht!]
\centering
\setlength{\tabcolsep}{2pt}
{\footnotesize
\begin{tabular}{ccccccccccccc}
\toprule
\textbf{GPTQ} & \textbf{TFF} & \textbf{$2\sigma$ clip} & \textbf{Redundancy} & \multicolumn{3}{c}{\textbf{ViT}} & \multicolumn{3}{c}{\textbf{DeiT III}} & \multicolumn{3}{c}{\textbf{Swin}} \\
& & & $r = 1.1$ &  \textbf{S}& \textbf{B}& \textbf{H}& \textbf{S}& \textbf{B} & \textbf{H} & \textbf{S} & \textbf{B} & \textbf{L} \\
\midrule
{\color{green} ON} & {\color{red} OFF}  & {\color{red} OFF}     & {\color{red} OFF}     & $0.4 $    & $29.26$   & $42.63$   & $0.45 $   & $8.5  $   & $28.2 $   & $43.54$   & $47.38$  &  $71.69$ \\
{\color{green} ON} & {\color{green} ON} & {\color{red} OFF}     & {\color{red} OFF}     & $0.88$    & $59.87$   & $68.75$   & $1.48 $   & $29.92$   & $84.33$   & $61.01 $   & $60.21$  &  $79.52$ \\
{\color{green} ON} & {\color{green} ON} & {\color{green} ON}    & {\color{red} OFF}     & $48.10$    & $79.53$   & $85.49$   & $51.13$   & $77.99$   & $86.62$   & $77.91$   & $80.16$  &  $84.25$ \\
\rowcolor{Gray}
{\color{green} ON} & {\color{green} ON} & {\color{green} ON}    & {\color{green} ON}    & $\bm{61.51}$   & $\bm{80.93}$   & $\bm{85.99}$   & $\bm{65.33}$   & $\bm{80.91}$   & $\bm{86.68}$   & $\bm{78.77}$   & $\bm{81.33}$  &  $\bm{84.42}$ \\ 
\midrule
& & & Full Precision & $81.39$ & $85.1$ & $87.59$ & $83.06$ & $85.7$ & $87.19$ & $82.79$ & $84.7$ & $85.95$ \\
\bottomrule
\end{tabular}
}
\caption{Incremental impact of various steps in FrameQuant on ImageNet-1k accuracy for different Transformer models in Vision}
\vspace{-3pt}
\label{tab: ablation study}
\vspace{-0.1in}
\end{table*}

\textbf{Impact of Gaussian assumption on the weights distribution}. Figure \ref{fig:two-sigma} shows a
representative example of the distribution of weights in a model from the ViT family and why the $2\sigma$ clipping seems reasonable 
for capturing most of the mass. The weights distribution for models from DeiT and Swin Transformer are shown in Figure \S\ref{fig:two-sigma DeiT Swin}. 
\vspace{-5pt}


\subsection{Results on Language Models}
\label{sec:llm experiments}
\begin{figure}[!tb]
    \centering
    \begin{subfigure}{.495\linewidth}
        \centering
        \includegraphics[width=\linewidth]{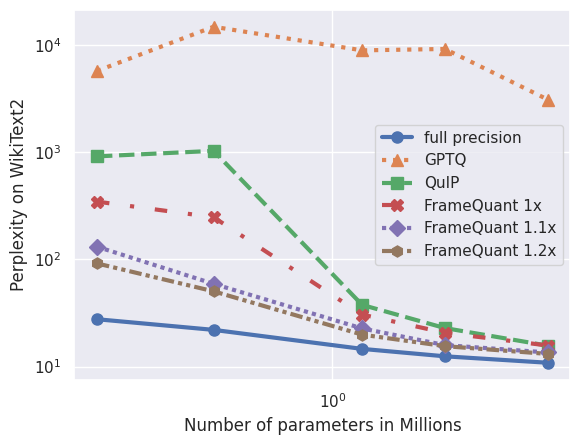}
        \caption{Perplexity on WikiText2}
        \label{fig:WikiText2 OPT}
    \end{subfigure}%
    \begin{subfigure}{.495\linewidth}
        \centering
        \includegraphics[width=\linewidth]{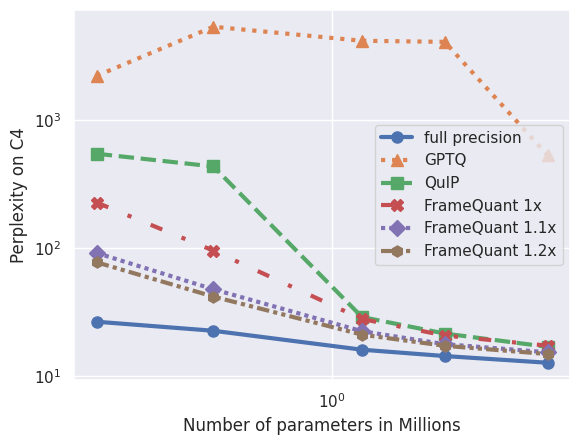}
        \caption{Perplexity on C4}
        \label{fig:C4 OPT}
    \end{subfigure}%
    \vspace{-0.1in}
    \caption{\label{fig:perplexity OPT}Perplexity of models from OPT family on WikiText2 and C4 datasets. FrameQuant performs better than all other quantization methods under consideration. We can also see that the performance gap between the quantized models and the unquantized model goes down as the size of the models increases.}
\vspace{-5pt}
\end{figure}

In this experiment, we evaluate the perplexity of quantized models from the OPT \cite{zhang2022opt} and Llama2 \cite{touvron2023llama} family on two datasets - WikiText2 \cite{merity2017pointer} and C4 \cite{JMLR:v21:20-074}. Figure \ref{fig:perplexity OPT} shows the perplexity of models from the OPT family as the size is increased. We see that FrameQuant at $1\times$ redundancy performs better than all other quantization methods. With a redundancy factor of $1.1\times$, FrameQuant reduces the performance gap with the full precision models as suggested by the theory. We see similar results for models from the Llama2 family as well. We also finetuned the Llama2-7B model quantized by various methods on diverse downstream tasks and observed a maximum accuracy boost of $41\%$ by FrameQuant at $r=1.1\times$ compared to vanilla GPTQ. Table \ref{tab:llm_wt2_ppl} summarizes the perplexity of all the models on the WikiText2 \cite{merity2017pointer} dataset. Results on downstream tasks/additional datasets is in  Appendix \S\ref{sec: more exp on llms}.

\subsection{Comparision with Mixed-precision Quantization}
A redundancy factor of $1.1$ is the same as an average bit-width of $2.2$ per weight parameter. Mixed-precision quantization methods can achieve fractional bit-widths by using different bit-widths for different weights in the model. We compare FrameQuant with a recent Mixed-precision method, ZeroQuant \cite{NEURIPS2022_adf7fa39}. We test FrameQuant with a bit-width of $2$ and a redundancy factor of $1.1$ relative to ZeroQuant at different fractional bit-widths. As shown in Table \ref{tab:zeroquant}. FrameQuant performs favorably with ZeroQuant, even at low bit widths.

\vspace{-10pt}

\begin{table}[!bt]
\centering
\setlength{\tabcolsep}{5pt}
{\footnotesize
\begin{tabular}{cccc}
\toprule
\textbf{Method} & \textbf{\#bits} & \textbf{acc} & \textbf{mm-acc} \\
\midrule
Full-Precision & 32 & $84.19$ & $84.67$ \\
\midrule
ZeroQuant & 4.33 & $78.69$ & $78.07$ \\
ZeroQuant & 3.66 & $54.91$ & $56.45$ \\
ZeroQuant & 2.66 & $38.00$ & $38.30$ \\
\rowcolor{Gray}
FrameQuant  & 2.2 & $\bm{80.02}$	& $\bm{79.37}$ \\
\bottomrule
\end{tabular}
}
\caption{Performance of the BERT model quantized with ZeroQuant and FrameQuant on the MNLI dataset. FrameQuant performs better than ZeroQuant even with a lower bit-width than ZeroQuant.}
\vspace{-7pt}
\label{tab:zeroquant}
\end{table}

\begin{table}[!tb]
\centering
\setlength{\tabcolsep}{2.5pt}
{\footnotesize
\begin{tabular}{>{}c>{}c>{}c>{}c>{}c>{}c|>{}c>{}c}
\toprule
\textbf{Method} & \textbf{bits} & \multicolumn{4}{c}{\textbf{OPT}} & \multicolumn{2}{c}{\textbf{Llama2}} \\
& & \textbf{125M}& \textbf{1.3B}& \textbf{2.7B}& \textbf{6.7B}& \textbf{7B}& \textbf{70B}\\
\midrule
Full-Precision & 16 & 27.65 & 14.62 & 12.47 & 10.86 & 5.68 & 3.32 \\
\midrule
GPTQ    & 2 & 5.7e3 & 8.9e3 & 9.1e3 & 3.1e3 & 6.4e3 & 140.5 \\
QuIP    & 2 & 913.0 & 37.59 & 22.86 & 15.67 & 26.02 & 6.21 \\
\rowcolor{Gray}
FrameQuant & 2 & 345.7 & 30.54 & 20.67 & 15.72 & 14.85 & 5.50\\
\rowcolor{Gray}
FrameQuant  & 2.2 & \textbf{131.2} & \textbf{22.68} & \textbf{15.86} & \textbf{13.53} & \textbf{8.48} & \textbf{4.67} \\
\bottomrule
\end{tabular}
}
\caption{Perplexity (lower is better) of Llama2 and OPT models on WikiText2 dataset when quantized to 2 (or 2.2) bits by different methods.}
\vspace{-15pt}
\label{tab:llm_wt2_ppl}
\end{table}

\section{Other Practical Considerations}
\subsection{Storage requirements}
Weight quantization has a direct improvement on the storage needs of the models. Table \ref{tab:storage sizes} shows the sizes of  compressed Llama2 models. FrameQuant reduces the size of the models by around $85\%$ on average.  

\begin{table}[!bt]
\centering
\setlength{\tabcolsep}{5pt}
{\footnotesize
\begin{tabular}{ccc}
\toprule
 & \textbf{Llama2 7B} & \textbf{Llama2 13B} \\
\midrule
Original model & $13$G & $25$G \\
FrameQuant & $2.1$G & $3.6$G \\
\bottomrule
\end{tabular}
}
\caption{Size of  original and quantized model with FrameQuant.}
\vspace{-10pt}
\label{tab:storage sizes}
\end{table}

\subsection{Inference speeds}
Since FrameQuant involves additional operations to compute and transform the weights from the low-bit Fusion Frame representations to the regular weight space, the raw inference speed is expected to be lower than GPTQ. On the other hand, at ~2 bits, the accuracy/perplexity of FrameQuant is much better than GPTQ. So, there is a trade-off. Table \ref{tab:inference speeds} shows the inference speeds of the quantized models on a Nvidia A100 GPU. Here, we used the block diagonal structure of Fusion Frames and a Hadamard transform-based fast random projection based on \cite{daoailab,10.5555/3618408.3620114} for the rotation matrices. This inference speed can be improved by using efficient kernels to load the weights into the GPU and perform the transformations.

\begin{table}[!bt]
\centering
\setlength{\tabcolsep}{5pt}
{\footnotesize
\begin{tabular}{ccc}
\toprule
\textbf{Method} & \textbf{Llama2 7B} & \textbf{Llama2 13B} \\
\midrule
GPTQ & $1425.07$t/s & $844.03$t/s \\
FrameQuant & $974.20$t/s & $607.01$t/s \\
\bottomrule
\end{tabular}
}
\caption{Inference speed in tokens/sec (t/s) of quantized models with GPTQ and FrameQuant. }
\vspace{-15pt}
\label{tab:inference speeds}
\end{table}


{\section{Discussion}
We cover a few additional aspects that were not explicitly discussed thus far. 
\begin{inparaenum}[\bfseries (1)]
\item {\em Can we reduce to one bit?} We performed experiments with redundancy of $1.8\times$ with $1$ bit per weight but were unsuccessful. For one bit, once the redundancy 
has exceeded $r=2$, it makes more sense to just use two bits. 
\item {\em Can FrameQuant run as QuIP?} For each layer, if we choose a Fusion Frame with a redundancy factor $r= 1$ and 
the random orthonormal basis $P_l, P_{\rm prev}$, we 
get a setup similar to QuIP \cite{chee2023quip} after removing the $2 \sigma$ weight clipping.
This is also why when QuIP is augmented with our $2\sigma$ clipping we see similar results to FrameQuant with $1\times$ redundancy.
\item {\em Additional storage needed?}: Since there are 
efficient deterministic algorithms to generate Fusion Frames, 
during inference, only knowledge of $(k,\rho, d)$ is needed. For rotations, we only need knowledge of the seed. Also, since many layers in a Transformer model have the same shape, these parameters can be shared across layers. Additional details on the storage benefits are  in \S\ref{sec:storage benefits} 
\item {\em Why is flexibility useful?} 
If the performance hit at the two-bit level is unacceptable for 
an application, the 
only recourse currently is to move up to three bits for existing methods ($50\%$ increase). 
However, FrameQuant allows flexibility through the choice of 
the redundancy factor $r$. 
\item {\em Higher bitwidths?} The main focus of this work is to evaluate 2-bit quantization of the weights in Vision and Language models and to check the benefits of applying Fusion Frames in terms of flexible bit-widths. Higher bit widths such as 3 or 4-bit quantization have been studied \cite{frantar2023optq, chee2023quip} and also used in practice \cite{llama_cpp}.
\item {\em Computational complexity during Inference:} The core FF-related compute is similar to alternatives \cite{chee2023quip} with a small overhead related to the number of subspaces $k$. During inference, we need an additional compute of $\mathcal{O}(d^2(kr+\log{d}))$ for transforming the weights from the Fusion Frame representation space to the regular weight space. Any quantization scheme in the low-bit regime will incur a cost of $\mathcal{O}(d^2)$ to transform the quantized weights by scaling and shifting them. More details are provided in \S\ref{sec:computational complexity during inference}.
\item {\em Quantization aware training:} FrameQuant can be modified to be applicable during QAT although we do not include such experiments here. One option is to use it during fine-tuning where the quantization loss is simulated, which can then be used to regularize the loss to make it more robust to quantization. Fusion Frames can meaningfully inform this bias, via an estimate of the ``out of subspace error'' to minimize degradation due to quantization. 
\item {\em Scaling laws vis-\`{a}-vis FrameQuant?} 
During quantization, the number of parameters does not change. Instead, each parameter has a lower degree of freedom since the number of states it can represent is reduced. We can use the (number of parameters $\times$ bit-width) as a proxy for the degree of freedom for each (quantized) model. Taking the quantization bit width into account, a line plot of test loss (on the vertical-axis) as a function of (number of parameters $\times$ bit-width) on the horizontal axis may have a different slope compared to \cite{kaplan2020scaling}, Fig. 1.
\item  {\em Rationale for clipping:} Let $u$ be a vector in $p$ dimensions. Let $P$ be a projection onto a random subspace in $p'$ dimensions. Projecting $u$ using $P$ gives $v$ as $v = P u$. Assume that the entries in $u$ have finite mean and variance and are uncorrelated. Then each entry of $v$ is effectively a sum of many scaled random variables. The distribution of these entries (sum of scaled variables, suitably standardized) approaches a normal distribution as the dimensionality $p$ grows. Weak dependence or mixing can also 
be handled. 
\end{inparaenum}

\section{Conclusions}

This paper describes {\em FrameQuant}, a Frames based 
algorithm for flexible low-bit quantization. 
Quantization is motivated by the need to efficiently serve Large Language Models on heterogeneous devices and flexibility here means that while we retain 
the option to go as low as two bits; depending 
on the needs of the downstream task, the user 
also has the flexibility to seek models with a net footprint 
of 2.x bits on average.
Across most widely used Vision Transformer models and Large Language models, we find that effective quantization 
is possible with only a small loss in performance relative 
to the full-precision model. 
Further, flexibility for a minor increase in redundancy is available and uniformly helps close the gap with full precision models. 
We observe, consistent with the literature, that quantization to low bit width is more favorable for larger models (in terms of a performance hit) than a similar quantization applied to smaller models. 
While some benefits (e.g., model loading time, loading larger models) are immediate, 
tighter integration with the hardware can unlock 
far more efficiency gains. The code is publicly available. 

\section*{Impact Statement}
This paper introduces a low precision quantization method for inference. The objective is to decrease memory needs and facilitate the implementation of larger models on less powerful devices, thereby reducing costs (economic impact) and the carbon footprint (environmental impact). We have not identified any particular ethical issues that need to be emphasized.

\section*{Acknowledgments}
Support through the Google Cloud Platform provided the computational resources for conducting our experiments.  The research was also supported in part by a Google gift award to UW-Foundation and funding from the Vilas
Board of Trustees.

\bibliography{example_paper}
\bibliographystyle{icml2024}

\newpage
\appendix
\onecolumn

\section*{Appendix}

In this Appendix, we provide additional details related to the experiments reported in the main paper. This Appendix is organized as follows. In {\em Section \ref{sec:Impact of redundancy in representations}} we analyze the impact of redundancy on the performance of the model in terms of classification accuracy on the ImageNet-1K dataset. In {\em Section \ref{sec:redundancy attn maps}}, we study this effect on the performance of Vision Transformer models, evaluated using activation maps. 
Next, in {\em Section \ref{sec:Number of Calibration images}}, we study the effect of the size of the calibration data used for quantizing various Vision Models.
In {\em Section \ref{sec: 2 sigma}}, we analyze the choice of the $2\sigma$ threshold for clipping the weights during quantization. We provide empirical evidence for different classes of Vision models. We also show that $2\sigma$ clipping alone cannot improve quantization performance. On the contrary, it can degrade the performance for weight configurations that are poorly distributed.
{\em Section \ref{sec: deit swin weight distribution} }shows the distribution of weights in the DeiT and Swin Transformer models.
In {\em Section \ref{sec:Activation Quantization}}, we present a framework for quantizing activations and show how the FF representation of activations inherently addresses the key pain points described in previous works. We follow this with a simple experiment with activation quantization enabled.
 In {\em Section \ref{sec: seg and obj det}}, we provide experiments on Segmentation and Object detection tasks.
In {\em Section \ref{sec: more exp on llms}}, we present more experiments on Language models on different datasets and downstream tasks as mentioned in the main paper. 
Then, in {\em Section \ref{sec:Robustness guarantees}}, we provide an expanded synopsis of the theoretical results that apply to our setting, as briefly described in the main paper. 
In {\em Section \ref{sec:construct tff}} we provide a brief synopsis of the algorithm used to generate a TFF for the curious reader.
Finally in {\em Section \ref{sec:storage and comp complexity}} we give a detailed analysis of the storage benefits of FrameQuant and the computational complexity during inference.

\section{Impact of redundancy in representations}
\label{sec:Impact of redundancy in representations}
We consider the impact of redundancy in our Frame representation moving forward from 2 bits, incrementally increasing redundancy. Table \ref{tab:redundancy experiment} shows the performance of different models at different levels of redundancy. We observe that for large models, the original performance without any redundancy was already high, and adding redundancy did not impact their performance significantly. 
However, this is not the case for smaller models. Here, we see significant performance improvements (around $+21\%$ for the ViT-S model).

\begin{table*}[ht!]
\centering
\setlength{\tabcolsep}{5pt}
{\footnotesize
\begin{tabular}{ccccccccccc}
\toprule
\textbf{Redundancy} & \textbf{bits} & \multicolumn{3}{c}{\textbf{ViT}} & \multicolumn{3}{c}{\textbf{DeiT III}} & \multicolumn{3}{c}{\textbf{Swin}} \\
& & \textbf{S}& \textbf{B}& \textbf{H}& \textbf{S}& \textbf{B} & \textbf{H} & \textbf{S} & \textbf{B} & \textbf{L} \\
\midrule
$r = 1.00 $ & $(2.0 ~\text{bits})$      & $48.10 $   & $79.53$  & $85.49$   & $51.13$   & $77.99$   & $86.62$   & $77.91$   & $80.16$   & $84.25$ \\
$r = 1.05 $ & $(2.1 ~\text{bits})$      & $56.19$   & $79.97$  & $85.67$   & $58.74$   & $79.59$   & $86.58$   & $78.47$   & $80.41$   & $84.26$ \\
$r = 1.10 $ & $(2.2 ~\text{bits})$      & $61.51$   & $80.93$  & $85.99$   & $65.33$   & $80.91$   & $86.68$   & $78.77$   & $81.33$   & $84.42$ \\
$r = 1.15 $ & $(2.3 ~\text{bits})$      & $65.17$   & $81.27$  & $86.04$   & $69.54$   & $81.69$   & $86.67$   & $78.87$   & $81.88$   & $84.51$ \\
$r = 1.20 $ & $(2.4 ~\text{bits})$      & $66.53$   & $81.59$  & $86.11$   & $71.07$   & $81.98$   & $86.61$   & $79.56$   & $82.02$   & $84.56$ \\
$r = 1.25 $ & $(2.5 ~\text{bits})$      & $68.57$   & $81.74$  & $86.06$   & $73.48$   & $82.51$   & $86.55$   & $79.99$   & $82.26$   & $84.51$ \\
$r = 1.30$ & $(2.6 ~\text{bits})$      & $69.02$   & $81.77$  & $85.99$   & $74.40 $   & $82.54$   & $86.38$   & $79.92$   & $82.39$   & $84.65$ \\
\midrule
Full Precision & - & $81.39$ & $85.1$ & $87.59$ & $83.06$ & $85.7$ & $87.19$ & $82.79$ & $84.7$ & $85.95$ \\
\bottomrule
\end{tabular}
}
\caption{Performance of various quantized models on ImageNet-1K classification task as the redundancy in FrameQuant is increased. We see that increasing the redundancy closes the gap between the performance of the quantized model and the Full precision model}
\label{tab:redundancy experiment}
\end{table*}

\section{Does redundancy impact attention maps?}
\label{sec:redundancy attn maps}
In the main paper, we discussed how the performance of the models improves as we increase the redundancy in the Fusion Frames during quantization. In this section, we provide additional details on how redundancy affects the attention maps of Vision Transformers from different classes. We will focus mainly on the small and base models where we see significant improvement in the validation accuracy on ImageNet, as we increase the redundancy. Figures \ref{fig:attn_map_redundancy_vits}, \ref{fig:attn_map_redundancy_deits} and \ref{fig:attn_map_redundancy_deitb} show the attention maps of Vit-S, DeiT III -S, and Deit III - B models respectively. These models show an improvement in the range of $4.55\%$ to $23.27\%$ as we increase the redundancy from $r=1$ to $r=1.3$. This is reflected in the attention maps as well. We see that as the redundancy is increased, the attention regions concentrate around the objects of interest systematically.  This is consistent with the improvement in accuracy and can also be seen in Figure \ref{fig:acc_vs_red_small}.

\begin{figure*}[!bt]
    \centering
    \includegraphics[width=0.76\linewidth]{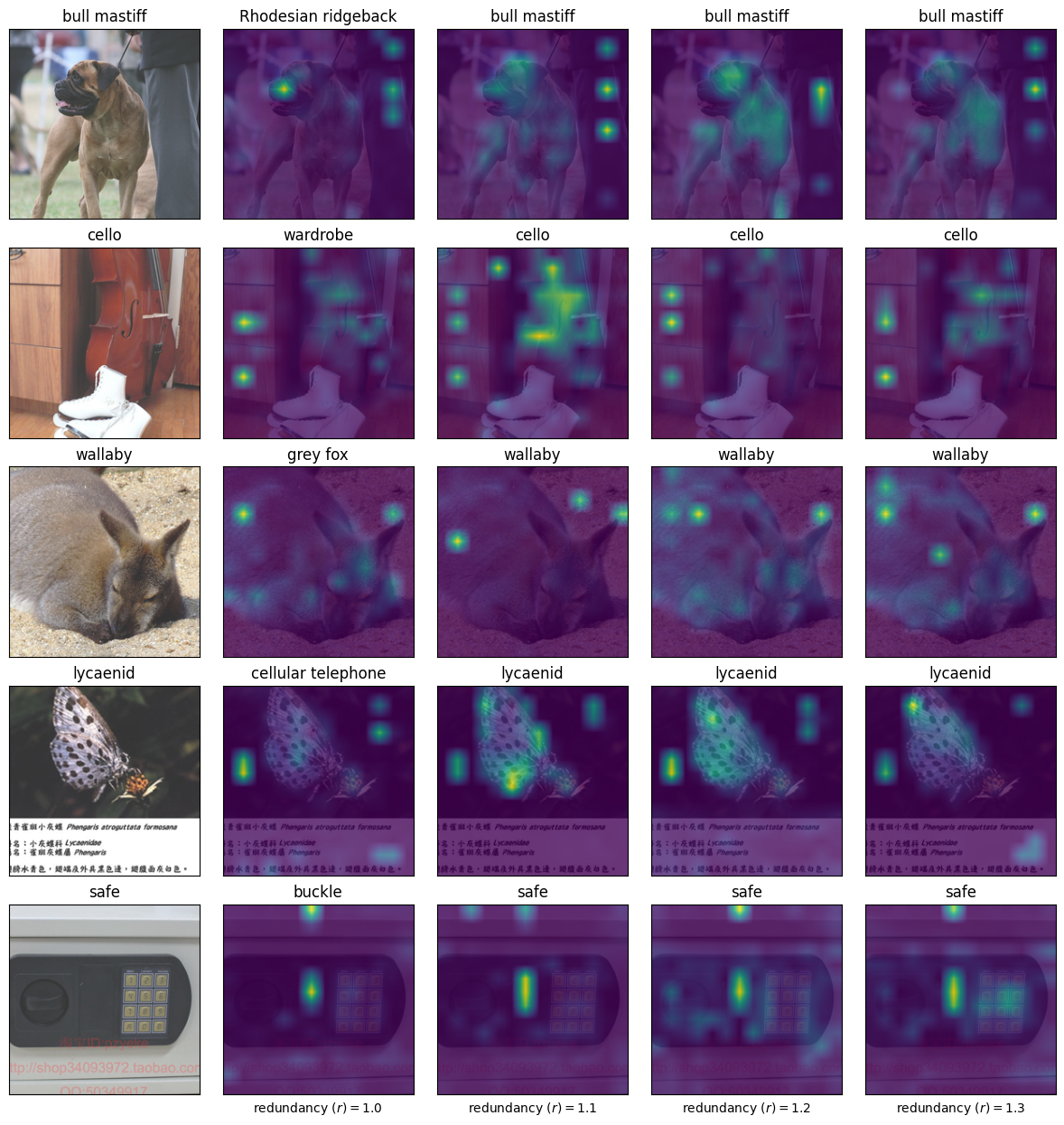}
    \vspace{-10pt}
    \caption{{\bf Effect of flexibility/redundancy on activation maps for ViT-S.} Figure showing attention maps of ViT-S as the redundancy is increased from $r=1$ to $r=1.3$ in the increments of $0.1$ from left to right. The first column shows the image and the ground truth label, and the rest of the columns show the regions that the model is attending to in the final transformer block. We see that as the redundancy is increased, the model gets more focused, with the attention regions concentrating on the objects of interest.}
    \label{fig:attn_map_redundancy_vits}
\end{figure*}

\begin{figure*}[!tb]
    \centering
    \includegraphics[width=0.76\linewidth]{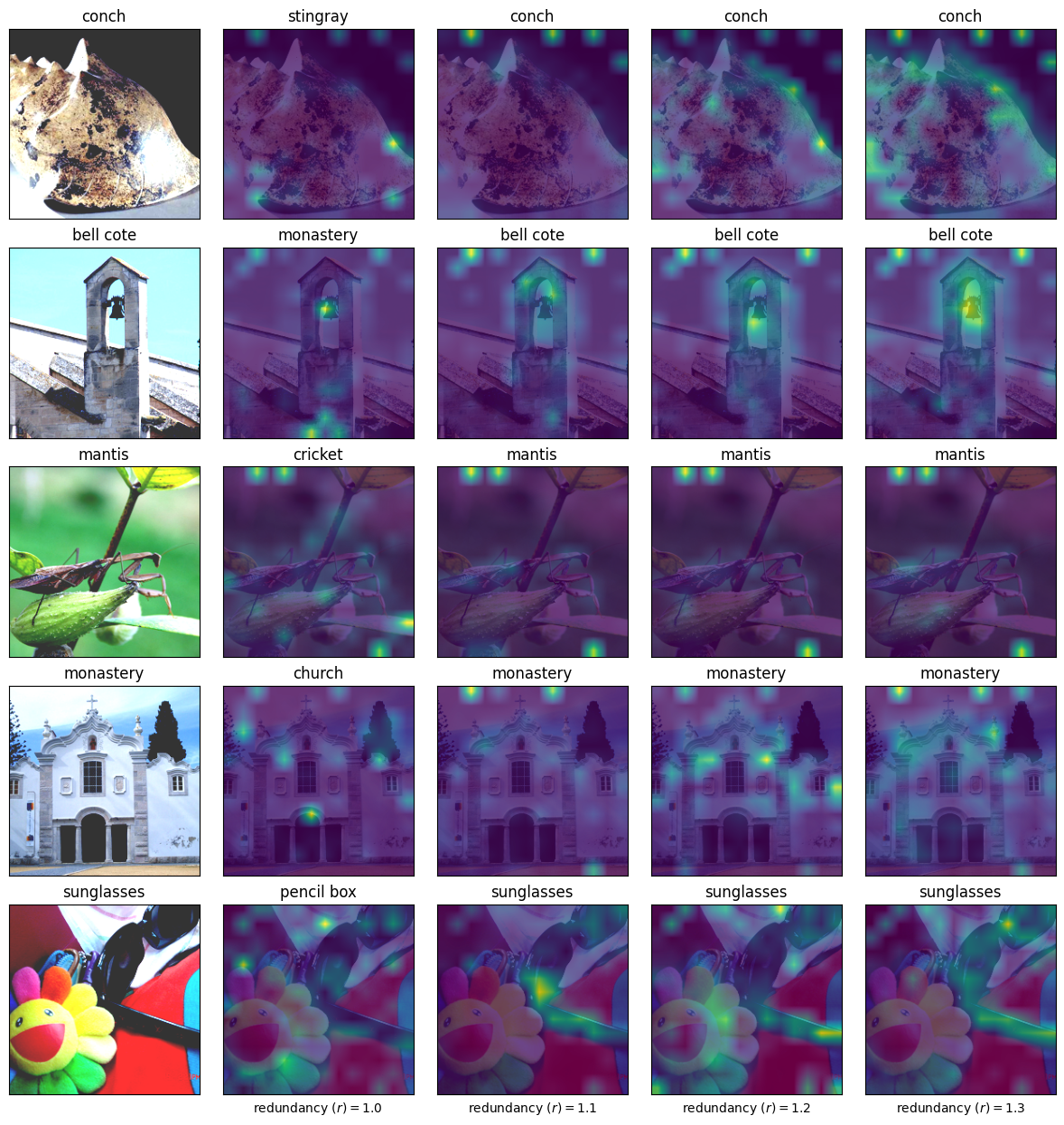}
    \vspace{-10pt}
    \caption{{\bf Effect of flexibility/redundancy on activation maps for DeiT III-S.} Figure showing attention maps of DeiT III -S as the redundancy is increased from $r=1$ to $r=1.3$ in the increments of $0.1$ from left to right.}
    \label{fig:attn_map_redundancy_deits}
\end{figure*}

\begin{figure*}[!tb]
    \centering
    \includegraphics[width=0.76\linewidth]{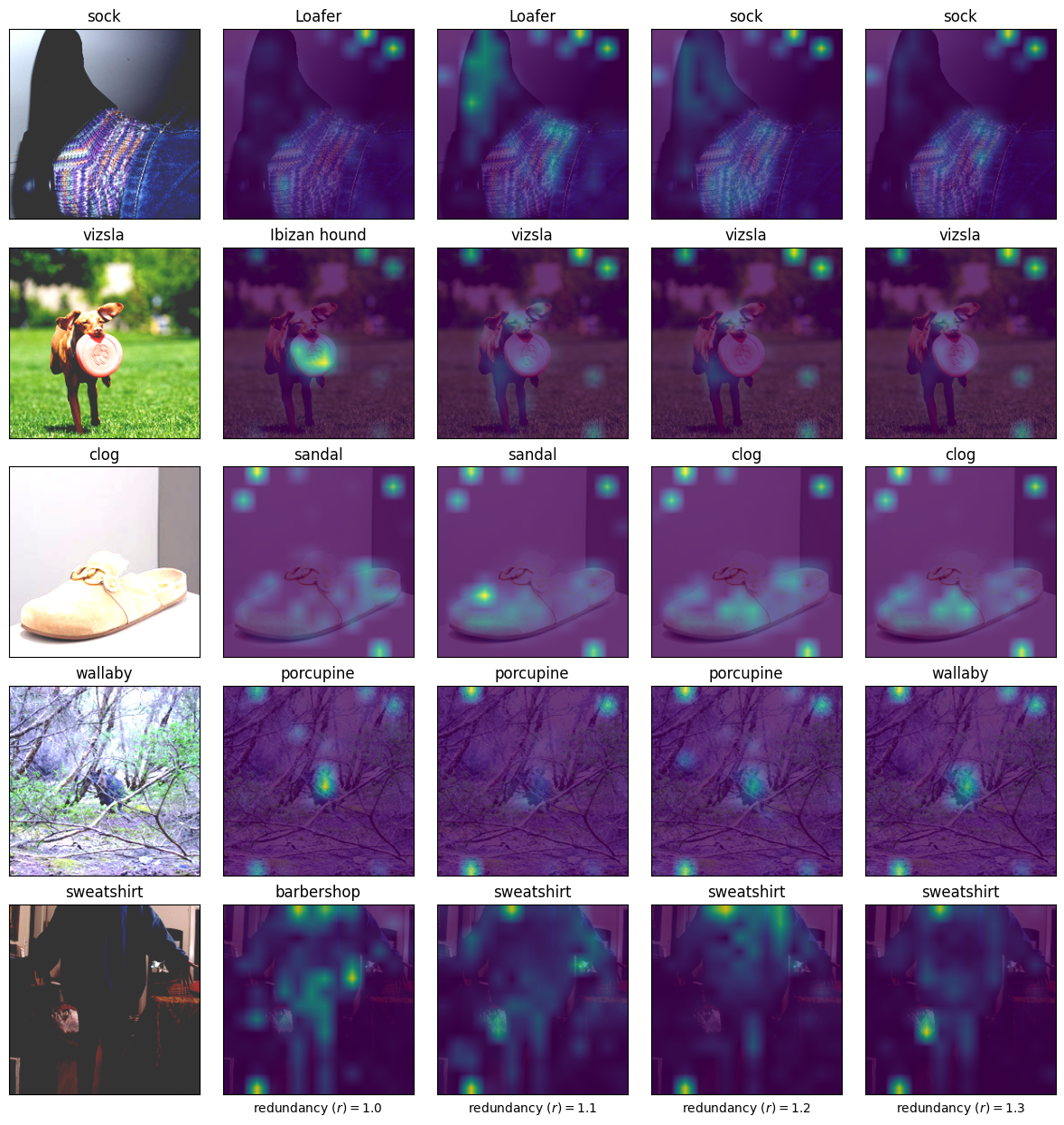}
    \vspace{-10pt}
    \caption{{\bf Effect of flexibility/redundancy on activation maps for DeiT III-B.} Figure showing attention maps of DeiT III -B as the redundancy is increased from $r=1$ to $r=1.3$ in the increments of $0.1$ from left to right.}
    \label{fig:attn_map_redundancy_deitb}
\end{figure*}

\begin{figure}[tb!]
    \centering
    \includegraphics[width=0.4\linewidth]{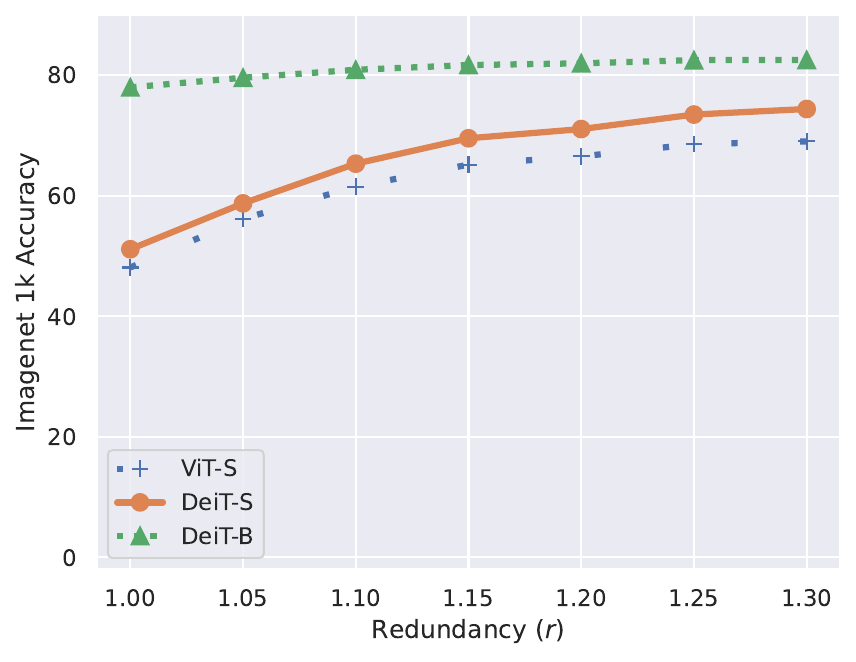}
    \caption{Trend of accuracies in small size models as we increase the redundancy}
    \label{fig:acc_vs_red_small}
\end{figure}

\begin{figure}[tb!]
    \centering
    \includegraphics[width=0.4\linewidth]{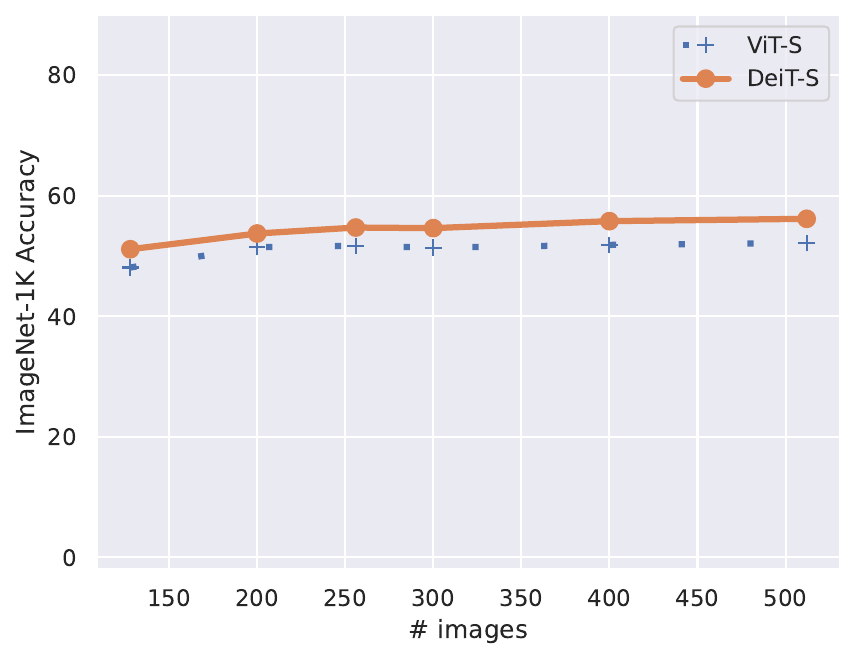}
    \caption{Trend of accuracies in small size models as we increase the number of calibration images}
    \label{fig:num_cal_images_fig}
\end{figure}

\begin{table*}[ht!]
\centering
\setlength{\tabcolsep}{5pt}
\begin{tabular}{cccccccccc}
\toprule
\textbf{\#images} & \multicolumn{3}{c}{\textbf{ViT}} & \multicolumn{3}{c}{\textbf{DeiT III}} & \multicolumn{3}{c}{\textbf{Swin}} \\
& \textbf{S}& \textbf{B}& \textbf{H}& \textbf{S}& \textbf{B} &\textbf{H}& \textbf{S} & \textbf{B}& \textbf{L} \\
\midrule
 $128$ & $48.10$ & $79.53$ & $85.49$ & $51.13$ & $77.99$ & $86.62$ & $77.91$ & $80.16$ & $84.25$ \\
  $200$ & $51.48$ & $79.84$ & $85.62$ & $53.74$ & $78.38$ & $86.61$ & $77.66$ & $80.19$ & $84.09$ \\
  $256$ & $51.69$ & $79.84$ & $85.74$ & $54.73$ & $79.06$ & $86.47$ & $77.96$ & $80.68$ & $84.31$ \\
\bottomrule
\end{tabular}
\caption{ImageNet-1K Top-1 validation accuracies of models from different classes as the number of calibration images is increased.}
\label{tab:num calib images exp}
\end{table*}

\section{Does the calibration set size matter?}
\label{sec:Number of Calibration images}
In the main paper, we noted that a small calibration set size was sufficient. In this section, we report on experiments varying the number of calibration images and observe the performance of different classes of models on ImageNet-1K. We use a redundancy factor of $r=1$ in this experiment. Table \ref{tab:num calib images exp} shows the validation accuracies for different classes of models as the number of calibration images is increased from $128$ to $256$. We can see that the performance improvement is seen only in the small-sized models from the ViT and DeiT III classes. So, we will focus on reporting results for these models. Figure \ref{fig:num_cal_images_fig} shows the accuracies of ViT-S and DeiT III-S models as the number of calibration images is increased from $128$ to $512$. We can see that there is a small improvement as the number of images is increased from $128$ to $200$, but the benefits taper off quickly as we increase it further. This shows that if access to the calibration is not limited, a small increase in the number of images used for quantization can benefit the final accuracies of the models, especially for smaller models.

\section{How does $2\sigma$ clipping affect performance?}
\label{sec: 2 sigma}
In the main paper, we discussed 
a simple clipping threshold at the $2\sigma$ level. 
In this section, we analyze the benefits of this choice and its effect on the performance of different classes of models on ImageNet-1K. As in the previous section, we use a redundancy factor of $r=1$ for these experiments and focus on the impact of clipping the weights at different levels based on their distribution. Figure \ref{fig:ACC vs sig} shows the accuracies of different classes of models as the threshold for the weights is varied from $\pm \sigma$ to $\pm 3 \sigma$. We can see that the performance of all the models peaks in the neighborhood of $\pm 2 \sigma$. Clipping at $\pm \sigma$ restricts the range of the weights too aggressively, incurring errors. At $\pm 3\sigma$ level, which is close to allowing the entire range, we are stretching the effective scale of the weights to allow all the extreme entries to be represented within the range. This, in turn, increases the width of the quantization levels, which affects the majority of the weights impacting performance. $\pm 2 \sigma$ seems to be the sweet spot.

\subsection{Does $2\sigma$ clipping alone improve performance?}
\label{sec:gptq+clip}
From our alation study \ref{sec: ablation study}, it might seem that $2\sigma$ clipping is doing the heavy lift in improving the performance. However, clipping is most effective once the weights are nicely distributed. A direct application of clipping on the weights has limited efficacy and incurs errors for weight configurations that are degenerate/poorly distributed. Projecting onto a space-filling basis makes clipping effective. To demonstrate this point quantitatively, we run GPTQ on Llama2 models with the $2\sigma$ clipping applied directly to the weights. Table \ref{tab:gptq+clip} shows that the performance degrades when the weights are clipped instead of their Fusion Frame representations as in FrameQuant.

\begin{table*}[bt!]
\centering
\setlength{\tabcolsep}{5pt}
{\footnotesize
\begin{tabular}{cccc}
\toprule
\textbf{Model} & \textbf{Quantization method} & \textbf{WikiText2} & \textbf{C4} \\
\midrule
Llama2 7B & GPTQ without clipping & 6.40e3 & 2.27e3 \\
Llama2 7B & GPTQ with clipping & 9.45e3 & 7.40e3 \\
Llama2 7B & FrameQuant with clipping & 14.85 & 19.62 \\
\midrule
Llama2 70B & GPTQ without clipping & 140.5 & 68.83 \\
Llama2 70B & GPTQ with clipping & 2.08e3 & 1.12e3 \\
Llama2 70B & FrameQuant with clipping & 5.5 & 7.85 \\
\bottomrule
\end{tabular}
}
\caption{Table showing the impact of clipping on GPTQ. FrameQuant computes the FF representations of the weights that are nicely distributed and can take advantage of clipping to remove outliers.}
\label{tab:gptq+clip}
\end{table*}

\begin{figure*}[!tb]
    \centering
    \begin{subfigure}{.32\textwidth}
        \centering
        \includegraphics[width=\linewidth,,height=0.8\linewidth]{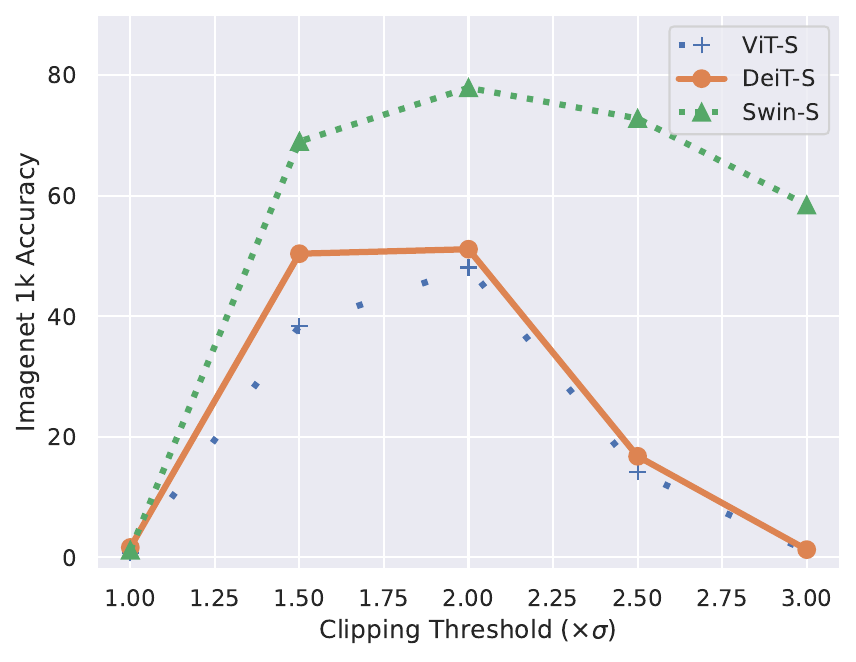}
        \caption{Accuracies of Small models}
        \label{fig:small vs sig}
    \end{subfigure}%
    \begin{subfigure}{.32\textwidth}
        \centering
        \includegraphics[width=\linewidth, ,height=0.8\linewidth]{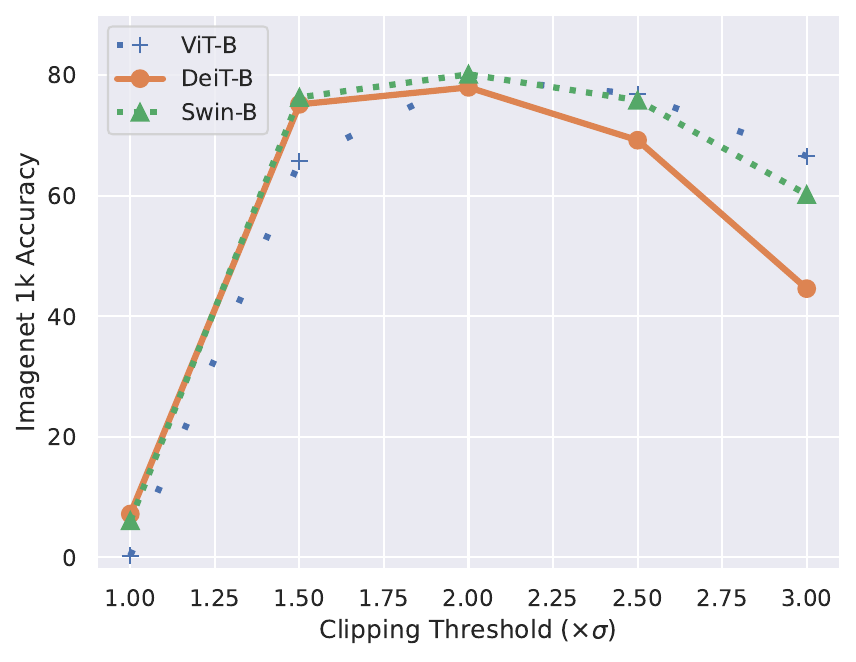}
        \caption{Accuracies of Base models}
        \label{fig:base vs sig}
    \end{subfigure}%
    \begin{subfigure}{.32\textwidth}
        \centering
        \includegraphics[width=\linewidth,height=0.8\linewidth]{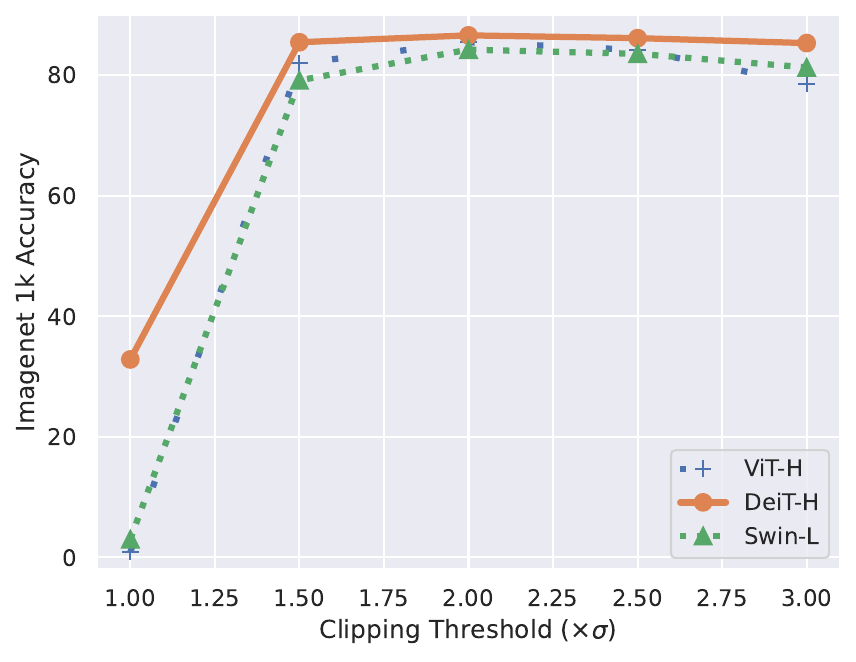}
        \caption{Accuracies of Large models}
        \label{fig:large vs acc}
    \end{subfigure}%
    \caption{Figure showing the impact of clipping at different thresholds based on $\sigma$}
    \label{fig:ACC vs sig}
    \vspace{-11pt}
\end{figure*}

\begin{figure}[!tb]
    \centering
    \begin{subfigure}{.48\textwidth}
        \centering
        \includegraphics[width=\linewidth]{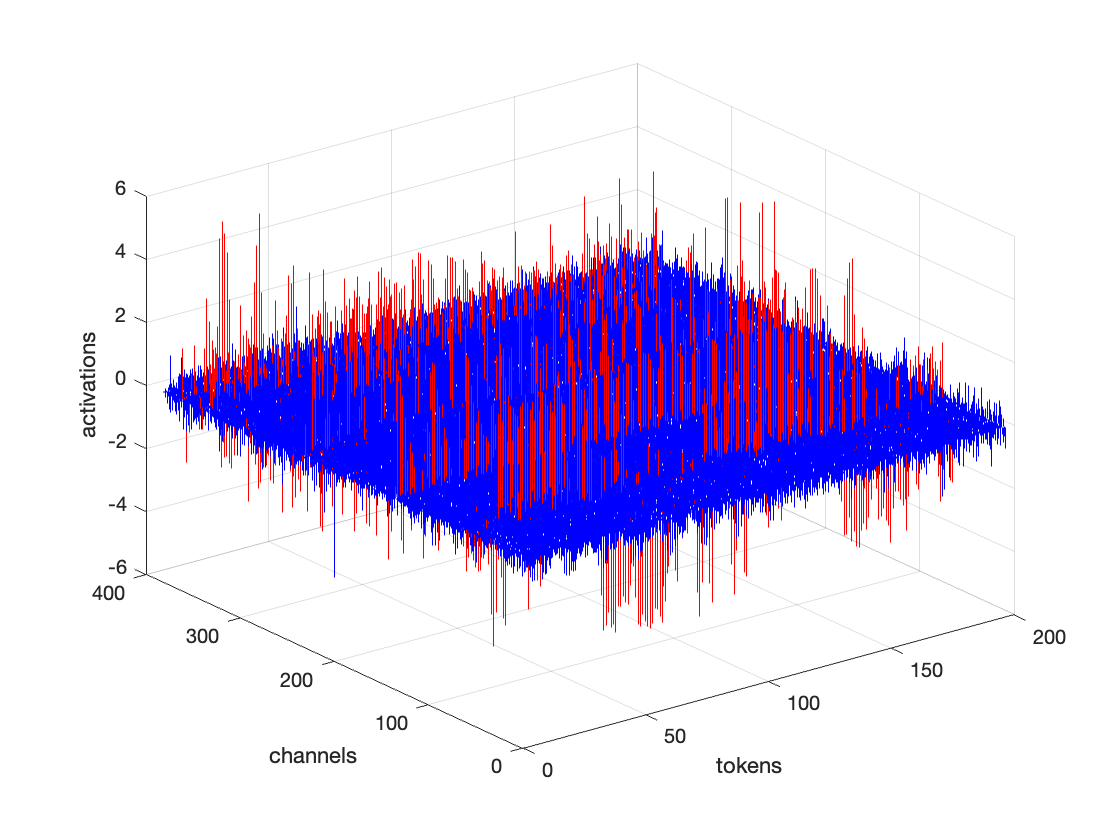}
        \caption{Activations of the first block in ViT-M}
        \label{fig:block1 vit-m activations}
    \end{subfigure}%
    \begin{subfigure}{.48\textwidth}
        \centering
        \includegraphics[width=\linewidth]{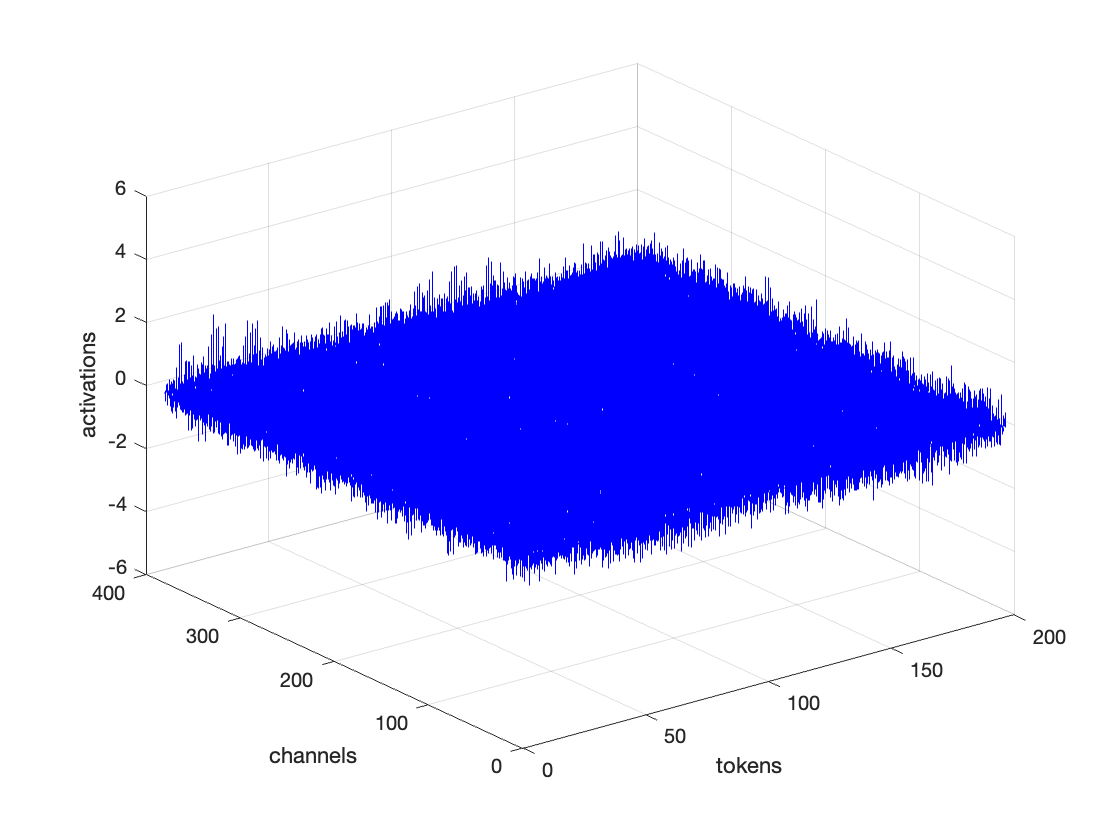}
        \caption{Activation FF representations of the first block in ViT-M}
        \label{fig:block1 vit-m activation FF represenatations}
    \end{subfigure}%
    \caption{Activations of the first Transformer Block of Vit-M model and their FF representations. We can see the outliers in the activations (shown in red) on the left, while the FF representations are well-behaved.}
    \label{fig:activations and their FF represenataions}
\vspace{-11pt}
\end{figure}

\section{Distribution of weights in the DeiT and Swin Transformer models}
\label{sec: deit swin weight distribution}
This section presents the distribution of the weights in the DeiT and Swin Transformer models. Figure \ref{fig:two-sigma DeiT Swin} shows the distribution of weights in a linear layer from the DeiT and Swin Transformer families. We can see that the distribution is well behaved and the $2\sigma$
threshold captures most of the mass well.

\begin{figure*}[!ht]
    \centering
    \begin{subfigure}{.32\textwidth}
        \centering
        \includegraphics[width=\linewidth]{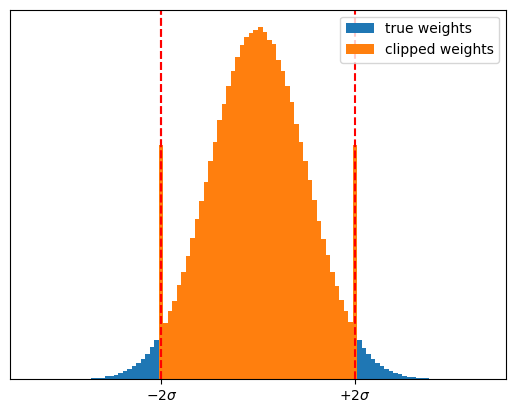}
        \caption{Weight distribution in DeiT}
        \label{fig:whist deit}
    \end{subfigure}%
    \begin{subfigure}{.32\textwidth}
        \centering
        \includegraphics[width=\linewidth]{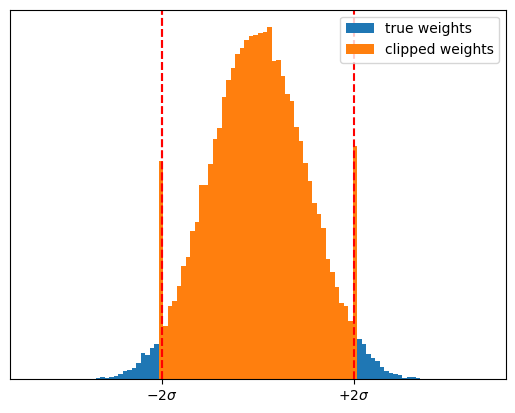}
        \caption{Weight distribution in Swin Transformer}
        \label{fig:whist swin}
    \end{subfigure}%
    \vspace{-5pt}
    \caption{Weights distribution in DeiT and Swin Transformer models.}
    \label{fig:two-sigma DeiT Swin}
\vspace{-11pt}
\end{figure*}

\section{FrameQuant for Activation Quantization?}
\label{sec:Activation Quantization}
In the main paper, we restricted the experimental setup of Vision Transformer models to 
weight quantization 
for meaningful comparisons to 
recent PTQ papers. 
This is because activation quantization 
in this low-bit regime has not been reported and 
each baseline will need modifications 
to report the best possible results. 
In this section, we provide some details regarding 
applying FrameQuant for activation Quantization with 
the caveat that a comprehensive head-to-head comparison to all reported baselines is difficult for the reasons above. 

{\bf Rounding activations to the nearest.} For smaller Transformer models, the inference efficiency bottleneck also largely lies in activations. So, we focus on these models to consider activation quantization. We performed activation quantization on ViT-S/B models with a simple rounding to the nearest, and we found that even when the weights are quantized to 2 (or 2.2) bits using FrameQuant, the performance drops are not large. This is promising and shows that FrameQuant is robust in preserving activations even at a 2.x bit level for weights. Table \ref{tab:simple act quant} shows the ImageNet-1K accuracy at different bit-widths for weights and activations.

\begin{table*}[ht!]
\centering
\setlength{\tabcolsep}{5pt}
{\footnotesize
\begin{tabular}{cccc}
\toprule
\textbf{Method} & \textbf{bits} & {\textbf{ViT-S}} & {\textbf{ViT-B}} \\
\midrule
Full Precision & W32/A32 & $81.39$   & $85.10$ \\
\midrule
FrameQuant & W2/A32 & $48.17$   & $79.53$ \\
FrameQuant & W2.2/A32 & $61.51$   & $80.93$ \\
\midrule
FrameQuant & W2/A8 & $48.02$   & $79.51$ \\
FrameQuant & W2.2/A8 & $60.96$   & $80.64$ \\
\midrule
FrameQuant & W2/A6 & $47.41$   & $78.59$ \\
FrameQuant & W2.2/A6 & $58.35$   & $80.14$ \\
\bottomrule
\end{tabular}
}
\caption{Performance of quantized ViT-S and ViT-B models on ImageNet-1K validation set. We used FrameQuant to quantize the weights while the activations are rounded to the nearest.}
\label{tab:simple act quant}
\end{table*}

{\bf Benefits of well-behaved FF representations.} 
Since we operate in the FF representation space, we can first compute the FF representations of the previous layer activations,
\begin{equation}
\textcolor{Green}{C_{\rm prev}} = \textcolor{Green}{P_{\rm prev}^T A_{\rm prev}}    
\end{equation}
and quantize these directly. 
Also, since activation quantization happens dynamically, during inference time, we keep the activation quantization procedure simple and 
just use the nearest rounding method. This can be written as:
\begin{equation}
    \textcolor{Green}{\bar{C}_{\rm prev}} = \nint{ \frac{\textcolor{Green}{C_{\rm prev}}}{\Delta_C} },  \quad \Delta_C = \frac{\max{|\textcolor{Green}{C_{\rm prev}}|}}{2^{\rm N-1}-1}
\end{equation}
where $\textcolor{Green}{\bar{C}_{\rm prev}}$ is in INT8 form and is the quantized version of the 
FF representations of the activations ($\textcolor{Green}{C_{\rm prev}}$). $\nint{\cdot}$ represents nearest rounding. We can substitute with $\lfloor \cdot \rfloor$ or $\lceil \cdot \rceil$ to get the floor or the ceil operation. 

As noted by \cite{xiao2023smoothquant}, we also observe that the activations have large outliers in some of the channels whose values are more than $100\times$ larger than the activations of other channels on average and this behavior is consistent across the tokens. This is shown in Figure \ref{fig:block1 vit-m activations}. So, to quantize the outliers, we need a large-scale factor $\Delta_C$, which will quantize all small values to zero. The other option is to use per-channel quantization -- where we have different scale factors for different channels. This would solve the outlier problem, but it is not ideal because we cannot use integer kernels for matrix multiplications in the Linear Layers. To use integer arithmetic for the matrix multiplications in the Linear layers, we can only perform per-token quantization for the activations and per-channel quantization for the weights. To solve this problem, \cite{xiao2023smoothquant} shifts the scale from activations to weights that are well-behaved. They dynamically search for different amounts of shifts between the weights and activations using a calibration set and use that during inference. Since we operate in the FF representation space, we observe that after we compute the FF representations of the activations, they are well-behaved. Figure \ref{fig:block1 vit-m activation FF represenatations} shows the FF representation of activation of the first Transformer block in the ViT-M model. So, we do not need to perform further scaling to reduce the range. This makes FrameQuant to be amenable to activation quantization if necessary in practice.

\section{Quantizing Segmentation and Object Detection models}
\label{sec: seg and obj det}
We used FrameQuant to quantize the Swin backbone for Object Detection and Segmentation Models. We compare our results with RepQ-ViT \cite{li2023repq}, one of the state-of-the-art publicly available quantization methods in this regime. Since our primary focus is quantizing the weights of the Transformer, for a fair comparison, we use RepQ-ViT to quantize the rest of the parameters, such as activations and norm layers. From Table \ref{tab:seg, obj det}, we can see that FrameQuant performs similarly to RepQ-ViT, and the main benefits of frameQuant kick in at very low bit widths.

\begin{table*}[bt!]
\centering
\setlength{\tabcolsep}{5pt}
{\footnotesize
\begin{tabular}{p{2cm} p{1.35cm} p{1.55cm} p{2.3cm} p{2.9cm}}
\toprule
\textbf{Method} & \textbf{Precision of Swin Backbone} & {\textbf{Precision of rest of the network}} & {\textbf{MoBY Mask RCNN w. Swin-T ($\text{AP}^{\text{box}}$ / $\text{AP}^{\text{mask}}$)}} & {\textbf{MoBY Cascade Mask RCNN with Swin-T ($\text{AP}^{\text{box}}$ / $\text{AP}^{\text{mask}}$)}} \\
\midrule
Full Precision & W32/A32 & W32/A32 & $43.6/39.6$   & $48.1/41.5$ \\
\midrule
RepQ-ViT & W6/A6 & W6/A6 & $42.6/39.0$ & $47.7/41.3$ \\
FrameQuant & W6/A6 & W6/A6 & $42.7/39.0$ & $47.8/41.3$ \\
\midrule
RepQ-ViT & W4/A4 & W4/A4 & $34.2/32.3$ & $43.8/38.6$ \\
FrameQuant & W4/A4 & W4/A4 & $34.5/32.5$ & $44.3/39.1$ \\
\midrule
RepQ-ViT & W3/A4 & W4/A4 & $27.5/26.4$ & $38.9/34.8$ \\
FrameQuant & W3/A4 & W4/A4 & $29.3/27.9$ & $41.2/36.7$ \\
\midrule
RepQ-ViT & W3/A4 & W3/A4 & $16.9/16.9$ & $32.4/29.2$ \\
FrameQuant & W3/A4 & W3/A4 & $21.7/21.5$ & $35.2/31.4$ \\
\bottomrule
\end{tabular}
}
\caption{Performance of quantized models with Swin-T backbone on the Object Detection and Segmentation tasks. We can see that FrameQuant performs similarly to RepQ-Vit at higher bit widths. The main benefits of Frame representations kick in at very low bit-widths.}
\label{tab:seg, obj det}
\end{table*}

\section{Additional Experiments on Language models}
\label{sec: more exp on llms}
\subsection{Evaluation on the C4 dataset}
This section is a continuation of section \ref{sec:llm experiments}. Here, we present the perplexity of different models from OPT and Llama2 classes on the C4 \cite{JMLR:v21:20-074} dataset. Consistent with our previous experiments, we see that FrameQuant with $1\times$ the redundancy performs better than all the methods under consideration. With an additional redundancy of $r=1.1\times$, FrameQuant closes the gap between the full precision model across all the sizes from different families of Large Language Models. The results are shown in table \ref{tab:llm_c4_ppl}.

\begin{table*}[!tb]
\centering
\setlength{\tabcolsep}{5pt}
\begin{tabular}{ccccccc|cc}
\toprule
\textbf{Method} & \textbf{\#bits} & \multicolumn{5}{c}{\textbf{OPT}} & \multicolumn{2}{c}{\textbf{Llama2}} \\
& & \textbf{125M}& \textbf{350M}& \textbf{1.3B}& \textbf{2.7B}& \textbf{6.7B}& \textbf{7B}& \textbf{70B}\\
\midrule
Full-Precision & 16 & $26.56$ & $22.58$ & $16.07$ & $14.34$ & $12.71$ & $7.26$ & $5.71$ \\
\midrule
GPTQ    & 2 & $2203.89$ & $5325.65$ & $4139.91$ & $4058.41$ & $528.41$ & $2265.09$ & $68.83$ \\
QuIP    & 2 & $543.63$ & $432.56$ & $28.91$ & $21.49$ & $16.92$ & $26.61$ & $8.65$ \\
\rowcolor{Gray}
FrameQuant ($r = 1.0$) & 2 & $226.15$ & $95.38$ & $27.90$ & $20.74$ & $17.28$ & $19.62$ & $7.85$\\ 
\rowcolor{Gray}
FrameQuant ($r = 1.1$) & 2.2 & $\bm{91.29}$ & $\bm{47.62}$ & $\bm{22.39}$ & $\bm{17.75}$ & $\bm{15.33}$ & $\bm{11.23}$ & $\bm{6.86}$ \\
\bottomrule
\end{tabular}
\caption{Perplexity (smaller the better) of Llama2 and OPT models on C4 dataset when quantized to 2 (or 2.2) bits by different methods.}
\label{tab:llm_c4_ppl}
\end{table*}

\subsection{Perplexity of Quantized Llama2 7B}
Figure \ref{fig:perplexity llama2-7b} shows the perplexity of Llama2-7B model quantized by different quantization schemes. We see that FrameQuant with a redundancy of 1x already performs better than all other methods. With increasing redundancy, the performance becomes closer to the Full precision model.

\begin{figure*}[!ht]
    \centering
    \begin{subfigure}{.48\textwidth}
        \centering
        \includegraphics[width=\linewidth]{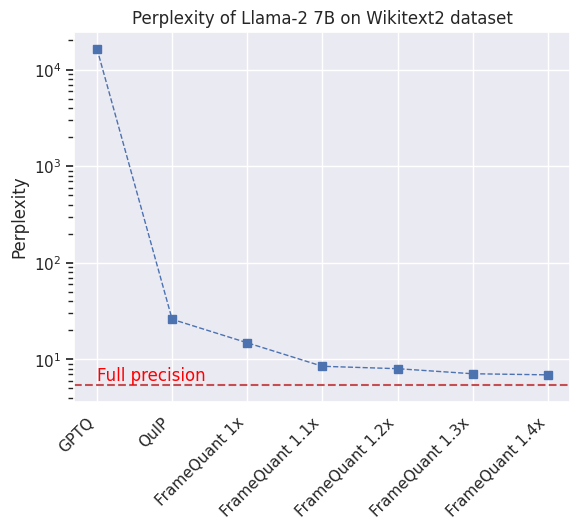}
        \caption{Perplexity of Llama2-7B model on WikiText2 dataset}
        \label{fig:WikiText2 llama2-7b}
    \end{subfigure}%
    \begin{subfigure}{.48\textwidth}
        \centering
        \includegraphics[width=\linewidth]{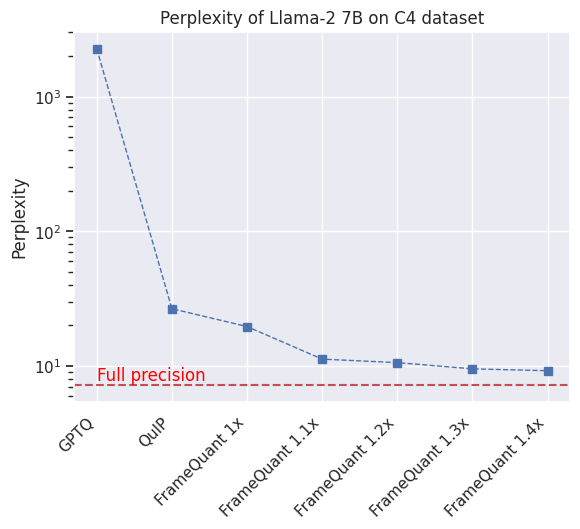}
        \caption{Perplexity of Llama2-7B model on C4 dataset}
        \label{fig:C4 llama2-7b}
    \end{subfigure}%
    \vspace{-5pt}
    \caption{Perplexity of Llama2-7B model on WikiText2 and C4 datasets. FrameQuant performs better than all quantization methods tested. With increasing redundancy, we see that the performance of the model also improves as indicated by the theory.}
    \label{fig:perplexity llama2-7b}
\vspace{-5pt}
\end{figure*}

\subsection{Performance on Downstream tasks}
In this experiment, we finetune the Llama2-7B model on downstream tasks. We ran experiments on ARC challenge, ARC easy \cite{clark2018think}, BoolQ \cite{clark2019boolq}, HellaSwag \cite{zellers2019hellaswag}, PIQA \cite{bisk2020piqa} and WinoGrande \cite{sakaguchi2021winogrande}. We used LM-evaluation harness \cite{eval-harness} for running our experiments on these diverse tasks. The results are presented in table \ref{tab:LM_eval harness}. We can see that again in line with our previous experiments, the LLM quantized with FrameQuant with no redundancy already performs better than all the other methods on the downstream tasks. With added redundancy, this performance goes up across all the tasks under consideration. Based on our previous experiments and as observed in \cite{chee2023quip}, we expect the performance gap between the full precision model and the quantized model to go down as the size of the models increases.

\begin{table*}[!tb]
\centering
\setlength{\tabcolsep}{4pt}
\begin{tabular}{cccccccc}
\toprule
\textbf{Method} & \textbf{\#bits} & \textbf{ARC (challenge)} & \textbf{ARC (easy)} & \textbf{BoolQ} & \textbf{HellaSwag} & \textbf{PIQA} & \textbf{WinoGrande} \\
\midrule
Full-Precision & 16 & $43.43$ & $76.35$ & $77.71$ & $57.16$ & $78.07$ & $69.06$ \\
\midrule
GPTQ    & 2 & $22.44$ & $24.58$ & $41.19$ & $25.93$ & $51.85$ & $50.43$ \\
QuIP    & 2 & $22.27$ & $42.76$ & $50.31$ & $34.04$ & $61.75$ & $52.64$ \\
\rowcolor{Gray}
FrameQuant ($r = 1.0$) & 2 & $23.98$ & $55.39$ & $63.52$ & $36.76$ & $66.65$ & $55.80$ \\
\rowcolor{Gray}
FrameQuant ($r = 1.1$) & 2.2 & $\textbf{31.91}$ & $\textbf{65.53}$ & $\textbf{67.95}$ & $\textbf{46.46}$ & $\textbf{73.07}$ & $\textbf{63.61}$ \\
\bottomrule
\end{tabular}
\caption{Evaluating Llama2-7B model quantized with different methods on a range of downstream tasks.}
\label{tab:LM_eval harness}
\end{table*}

\section{Robustness guarantees}
\label{sec:Robustness guarantees}
We provide additional details on two specific results (mentioned in the main paper) that apply to our construction. We encourage the interested reader
to refer to \cite{OleChristen5009,FFTaAPGCGKB2012} 
for a more comprehensive treatment of the topic. 
\\
\textbf{LMMSE estimation from fusion frame measurements.} For a given layer $l$ FrameQuant quantizes the transformed weights matrix $D_l$ which is given by $D_l = \textcolor{BlueViolet}{P_{l}^T} (\Theta_l \textcolor{BlueViolet}{P_{\rm prev}})$. We can treat $\hat{D}_l$ as a projection of $\Theta_l \textcolor{BlueViolet}{P_{\rm prev}}$ which is corrupted by noise. During inference, the activations of this layer are given by $Z_l = \textcolor{BlueViolet}{P_{l}} \hat{D}_l \textcolor{Green}{C_{\rm prev}}$. But, can we do better? Instead of directly applying the synthesis operator $\textcolor{BlueViolet}{P_{l}}$ to compute $Z_l$ from its FF representations $\hat{D}_l \textcolor{Green}{C_{\rm prev}}$, we can design a simple linear filter $F$ that minimizes the MSE in $Z_l$ because we are using a quantized $\hat{D}_l$. The final expression for the computation of the output of the layer will be $Z_l = F \hat{D}_l \textcolor{Green}{C_{\rm prev}}$. This linear MSE minimizer $F$ is known to be the {\em Wiener Filter} and has a closed-form expression with various levels of approximation. The following theorem states that the Wiener filter minimizes MSE when the Fusion Frame is {\em tight}.
\begin{theorem} \cite{KUTYNIOK200964}
    For the model described above, the MSE in linearly estimating the signal from its noisy projections is minimized when the Fusion Frame is tight
\end{theorem}
\textbf{Consistent Reconstruction.} Assuming the same mode of representing the modified weights $D_l$ as above, during inference, we can get a consistent estimate of the weights ($\hat{\Theta}_l$) from $\hat{D}_l$ if one were to solve a linear program for $\hat{X}$
\begin{equation*}
   \begin{bmatrix}
       \textcolor{BlueViolet}{P_{l}} \\
       \textcolor{BlueViolet}{-P_{l}}
   \end{bmatrix} \hat{X}_l \leq 
   \begin{bmatrix}
       \frac{\Delta}{2} + \hat{D}_l \\
       \frac{\Delta}{2} - \hat{D}_l
   \end{bmatrix},
\end{equation*}
where $\Delta$ is the quantization level. Here, the constraints in the Linear Program make sure that $\hat{X}$ belongs to the regions where valid unquantized values must lie, thereby removing the out-of-sub-space error \cite{650985}. We can get the estimated weights from $\hat{X}_l$ as $\hat{\Theta}_l = \hat{X}_l \textcolor{BlueViolet}{P^T_{\rm prev}}$. Using this consistent reconstruction yields estimates with an MSE which is upper bounded by $\mathcal{O}(1/r^2)$ \cite{650985}

\section{Synopsis of Construction of Tight Fusion Frames}
\label{sec:construct tff}
Here, we give a brief synopsis of an algorithm for generating Tight Fusion Frames for the curious reader. \cite{CASAZZA2011175} was the first to introduce a systematic method for constructing UNTFs (Unit Norm Tight Frames) that play a key role in constructing Tight Fusion Frames. They also characterize the $(k, \rho, d)$ values for which a Tight Fusion Frame exists. Whenever such a TFF exists, we can construct Tight Fusion Frames by using their algorithm. There are two main parts to the algorithm.
\begin{compactenum}
    \item Play Spectral Tetris to generate a UNTF of $d$ elements in $\mathbb{C}^\rho$
    \item Modulate this UNTF with complex roots of unity to generate a $(k,\rho,d)$ TFF for $\mathbb{C}^d$
\end{compactenum}
So, the first step is to generate a ``smaller'' frame and in the next step, we modulate the smaller frame to generate a ``larger'' Tight Fusion Frame. After generating a TFF for $\mathbb{C}^d$ we can easily extend it to the Real Field by applying the entrywise map $x + iy \mapsto \begin{bmatrix}
    x & -y \\
    y & x
\end{bmatrix}$. We describe the algorithm with the help of an example for the simplicity of explanation. We aim to construct a (5,4,11) TFF. So, $k=5, \rho=3, d=11$. 
\subsection{Spectral Tetris}
As the name suggests UNTFs are Tight frames where each frame vector has a unit norm. We construct a $4\times 11$ matrix $F$ whose columns are the frame vectors for $\mathbb{C}^4$ which satisfies
\begin{compactitem}
    \item Columns of unit norm 
    \item Orthogonal rows, meaning $FF^*$ is diagonal
    \item Rows of constant norm, meaning $FF^*$ is a constant multiple of identity matrix with the constant being $\frac{11}{4}$
\end{compactitem}
We start with a matrix \\
$$F = \begin{bmatrix}
    1 & 1 & ? & ? & ? & ? & ? & ? & ? & ? & ? \\
    ? & ? & ? & ? & ? & ? & ? & ? & ? & ? & ? \\
    ? & ? & ? & ? & ? & ? & ? & ? & ? & ? & ? \\
    ? & ? & ? & ? & ? & ? & ? & ? & ? & ? & ? \\
\end{bmatrix}$$\\
This leaves a norm of $\frac{11}{4}-2 = \frac{3}{4}$ to be filled in the first row. This can easily be added using a $2\times 2$ matrix $T(x)$ where $x = \frac{3}{4}$. $T(x)$ is defined as:
\begin{equation*}
    T(x) \vcentcolon  = \frac{1}{\sqrt{2}} \begin{bmatrix}
    \sqrt{x} & \sqrt{x} \\
    \sqrt{2-x} & -\sqrt{2-x}
\end{bmatrix}, \quad \quad T(x) T^*(x) = \begin{bmatrix}
    x & 0 \\
    0 & 2-x
\end{bmatrix}
\end{equation*}
After inserting $T(x)$, $F$ is now
$$F = \begin{bmatrix}
    1 & 1 & \frac{\sqrt{3}}{\sqrt{8}} & \frac{\sqrt{3}}{\sqrt{8}} & 0 & 0 & 0 & 0 & 0 & 0 & 0 \\
    0 & 0 & \frac{\sqrt{5}}{\sqrt{8}} & -\frac{\sqrt{5}}{\sqrt{8}} & ? & ? & ? & ? & ? & ? & ? \\
    0 & 0 & 0 & 0 & ? & ? & ? & ? & ? & ? & ? \\
    0 & 0 & 0 & 0 & ? & ? & ? & ? & ? & ? & ? \\
\end{bmatrix}$$\\
Then we continue adding ones in row two until the norm becomes less than $\frac{11}{4}$.
$$F = \begin{bmatrix}
    1 & 1 & \frac{\sqrt{3}}{\sqrt{8}} & \frac{\sqrt{3}}{\sqrt{8}} & 0 & 0 & 0 & 0 & 0 & 0 & 0 \\
    0 & 0 & \frac{\sqrt{5}}{\sqrt{8}} & -\frac{\sqrt{5}}{\sqrt{8}} & 1 & ? & ? & ? & ? & ? & ? \\
    0 & 0 & 0 & 0 & 0 & ? & ? & ? & ? & ? & ? \\
    0 & 0 & 0 & 0 & 0 & ? & ? & ? & ? & ? & ? \\
\end{bmatrix}$$\\
Now we insert $T(x)$ with the remaining norm. We repeat this process until all the rows are filled. The Final $F$ is given by
$$F = \begin{bmatrix}
    1 & 1 & \frac{\sqrt{3}}{\sqrt{8}} & \frac{\sqrt{3}}{\sqrt{8}} & 0 & 0 & 0 & 0 & 0 & 0 & 0 \\
    0 & 0 & \frac{\sqrt{5}}{\sqrt{8}} & -\frac{\sqrt{5}}{\sqrt{8}} & 1 & \frac{\sqrt{2}}{\sqrt{8}} & \frac{\sqrt{2}}{\sqrt{8}} & 0 & 0 & 0 & 0\\
    0 & 0 & 0 & 0 & 0 & \frac{\sqrt{6}}{\sqrt{8}} & -\frac{\sqrt{6}}{\sqrt{8}} & 1 & \frac{\sqrt{7}}{\sqrt{8}} & \frac{\sqrt{7}}{\sqrt{8}} & 0 \\
    0 & 0 & 0 & 0 & 0 & 0 & 0 & 0 & \frac{\sqrt{7}}{\sqrt{8}} & -\frac{\sqrt{7}}{\sqrt{8}} & 1 \\
\end{bmatrix}$$

\subsection{Modulation}
In the second step, we modulate the $F$ matrix with complex roots of unity, one subspace at a time. So, for each $k_i = 0,1,2,\dots k-1$, we construct a row vector 
$$w_{k_i} = \left[\left( e^{\frac{i2\pi k_i}{k}} \right) ^0 \left( e^{\frac{i2\pi k_i}{k}} \right) ^1 \left( e^{\frac{i2\pi k_i}{k}} \right) ^2 \dots \left( e^{\frac{i2\pi k_i}{k}} \right) ^{d-1}\right]$$
We multiply each row of $F$ with $w_{k_i}$ to generate the orthogonal basis for different subspaces indexed by $k_i$. Theorem $14$ by \citet{CASAZZA2011175} proves that the Fusion Frames generated by this algorithm are Tight. The Final Fusion Frame vectors are shown in Table \ref{tab:fusion frames constructed}.

\begin{table}[!bt]
    \centering
\begin{equation*}
\setlength{\arraycolsep}{0.9em}
\begin{bmatrix*}[l]
    1 & 1 & \frac{\sqrt{3}}{\sqrt{8}} & \frac{\sqrt{3}}{\sqrt{8}} & 0 & 0 & 0 & 0 & 0 & 0 & 0 \\
    1 & \omega & \frac{\sqrt{3}}{\sqrt{8}} \omega^2 & \frac{\sqrt{3}}{\sqrt{8}} \omega^3 & 0 & 0 & 0 & 0 & 0 & 0 & 0 \\
    1 & \omega^2 & \frac{\sqrt{3}}{\sqrt{8}} \omega^4 & \frac{\sqrt{3}}{\sqrt{8}} \omega & 0 & 0 & 0 & 0 & 0 & 0 & 0 \\
    1 & \omega^3 & \frac{\sqrt{3}}{\sqrt{8}} \omega & \frac{\sqrt{3}}{\sqrt{8}} \omega^4 & 0 & 0 & 0 & 0 & 0 & 0 & 0 \\
    1 & \omega^4 & \frac{\sqrt{3}}{\sqrt{8}} \omega^3 & \frac{\sqrt{3}}{\sqrt{8}} \omega^2 & 0 & 0 & 0 & 0 & 0 & 0 & 0 \\
    0 & 0 & \frac{\sqrt{5}}{\sqrt{8}} & -\frac{\sqrt{3}}{\sqrt{8}} & 1 & \frac{\sqrt{2}}{\sqrt{8}} & \frac{\sqrt{2}}{\sqrt{8}} & 0 & 0 & 0 & 0 \\
    0 & 0 & \frac{\sqrt{5}}{\sqrt{8}} \omega^2 & -\frac{\sqrt{3}}{\sqrt{8}} \omega^3 & \omega^4 & \frac{\sqrt{2}}{\sqrt{8}} & \frac{\sqrt{2}}{\sqrt{8}} \omega & 0 & 0 & 0 & 0 \\
    0 & 0 & \frac{\sqrt{5}}{\sqrt{8}} \omega^4 & -\frac{\sqrt{3}}{\sqrt{8}} \omega & \omega^3 & \frac{\sqrt{2}}{\sqrt{8}} & \frac{\sqrt{2}}{\sqrt{8}} \omega^2 & 0 & 0 & 0 & 0 \\
    0 & 0 & \frac{\sqrt{5}}{\sqrt{8}} \omega & -\frac{\sqrt{3}}{\sqrt{8}} \omega^4 & \omega^2 & \frac{\sqrt{2}}{\sqrt{8}} & \frac{\sqrt{2}}{\sqrt{8}} \omega^3 & 0 & 0 & 0 & 0 \\
    0 & 0 & \frac{\sqrt{5}}{\sqrt{8}} \omega^3 & -\frac{\sqrt{3}}{\sqrt{8}} \omega^2 & \omega & \frac{\sqrt{2}}{\sqrt{8}} & \frac{\sqrt{2}}{\sqrt{8}} \omega^4 & 0 & 0 & 0 & 0 \\
    0 & 0 & 0 & 0 & 0 & \frac{\sqrt{6}}{\sqrt{8}} & -\frac{\sqrt{6}}{\sqrt{8}} & 1 & \frac{\sqrt{7}}{\sqrt{8}} & \frac{\sqrt{7}}{\sqrt{8}} & 0 \\
    0 & 0 & 0 & 0 & 0 & \frac{\sqrt{6}}{\sqrt{8}} & -\frac{\sqrt{6}}{\sqrt{8}} \omega & \omega^2 & \frac{\sqrt{7}}{\sqrt{8}} \omega^3 & \frac{\sqrt{7}}{\sqrt{8}} \omega^4 & 0 \\
    0 & 0 & 0 & 0 & 0 & \frac{\sqrt{6}}{\sqrt{8}} & -\frac{\sqrt{6}}{\sqrt{8}} \omega^2 & \omega^4 & \frac{\sqrt{7}}{\sqrt{8}} \omega & \frac{\sqrt{7}}{\sqrt{8}} \omega^3 & 0 \\
    0 & 0 & 0 & 0 & 0 & \frac{\sqrt{6}}{\sqrt{8}} & -\frac{\sqrt{6}}{\sqrt{8}} \omega^3 & \omega & \frac{\sqrt{7}}{\sqrt{8}} \omega^4 & \frac{\sqrt{7}}{\sqrt{8}} \omega^2 & 0 \\
    0 & 0 & 0 & 0 & 0 & \frac{\sqrt{6}}{\sqrt{8}} & -\frac{\sqrt{6}}{\sqrt{8}} \omega^4 & \omega^3 & \frac{\sqrt{7}}{\sqrt{8}} \omega^2 & \frac{\sqrt{7}}{\sqrt{8}} \omega^1 & 0 \\
    0 & 0 & 0 & 0 & 0 & 0 & 0 & 0 & \frac{\sqrt{7}}{\sqrt{8}} & -\frac{\sqrt{7}}{\sqrt{8}} & 1 \\
    0 & 0 & 0 & 0 & 0 & 0 & 0 & 0 & \frac{\sqrt{7}}{\sqrt{8}} \omega^3 & -\frac{\sqrt{7}}{\sqrt{8}} \omega^4 & 1 \\
    0 & 0 & 0 & 0 & 0 & 0 & 0 & 0 & \frac{\sqrt{7}}{\sqrt{8}} \omega & -\frac{\sqrt{7}}{\sqrt{8}} \omega^3 & 1 \\
    0 & 0 & 0 & 0 & 0 & 0 & 0 & 0 & \frac{\sqrt{7}}{\sqrt{8}} \omega^4 & -\frac{\sqrt{7}}{\sqrt{8}} \omega^2 & 1 \\
    0 & 0 & 0 & 0 & 0 & 0 & 0 & 0 & \frac{\sqrt{7}}{\sqrt{8}} \omega^2 & -\frac{\sqrt{7}}{\sqrt{8}} \omega & 1 \\
    \end{bmatrix*}
\end{equation*}
    \caption{\textbf{$\bm{(5,4,11)}$-TFF} for $\mathbb{C}^{11}$. Here, $\omega = e^{i2\pi/5}$. Each pair of rows belongs to the same subspace if their indices differ by a multiple of 5}
    \label{tab:fusion frames constructed}
\end{table}


\section{Storage benefits and Computational complexity during inference}
\label{sec:storage and comp complexity}
\subsection{Storage benefits}
\label{sec:storage benefits}
 Consider an example where we are quantizing a weight matrix $\Theta_l$ of dimension $1024 \times 1024$ using FrameQuant with a redundancy factor of $r = 1.1 \times$. The size of the original matrix using FP32 is 4MB. After transforming the weights to map within the FF representation space, the transformed weights $D_l$ have dimensions $1126 \times 1126$, which are quantized and represented using 2 bits. This quantized weight $\hat{D}_l$ has a size of 0.3MB. Along with the quantized weights, we need to store the bias and scale values for each row leading to an additional storage of 1024 FP32 values, which will incur an additional cost of 0.007MB. All this sums up to a storage of 0.307MB from an initial 4MB giving a savings of 13x in the storage requirements. Since we can generate the Fusion Frames on the fly, we just need to store the $(k, \rho, d)$ values, and a seed to generate the random rotation matrix which incurs negligible storage costs. Table \ref{tab:storage sizes} shows the sizes of Llama2 models when compressed with FrameQuant.
 
\subsection{Computational Complexity during Inference}
\label{sec:computational complexity during inference}
 Consider a linear layer in a transformer model with weights $\Theta_l$ of dimensions $d \times d$. Using FrameQuant these weights are transformed to $D_l$ and the quantized weights $\hat{D}_l$ are stored. Let the parameters of the TFF used for quantization be $(k, \rho, d)$. As a recap, $k$ is the number of subspaces, $\rho$ is the dimension of each subspace and $d$ is the dimension of the Hilbert space we are operating in. So, the redundancy in Frame representations is $r = \frac{k \rho}{d}$. Let, $\textcolor{BlueViolet}{T_l}, \textcolor{BlueViolet}{T_{\rm prev}} \in \mathbb{R}^{d \times k\rho}$ be the vectorized Orthonormal basis for the current layer, and the previous layer respectively. During inference, the quantized weights $\hat{D}_l$ are transformed to the weight space as $\hat{\Theta}_l = \textcolor{BlueViolet}{P_l}  \hat{D}_l \textcolor{BlueViolet}{P_{\rm prev}^T} $. Here, $\textcolor{BlueViolet}{P_l} = R_l(\textcolor{BlueViolet}{T_l}), \textcolor{BlueViolet}{P_{\rm prev}} = R_{\rm prev}(\textcolor{BlueViolet}{T_{\rm prev}})$, where $R_l, R_{\rm prev} \in \mathbb{R}^{d \times d}$ denote the rotation matrices for the current and the previous layers respectively. So, the overall operation is $\hat{\Theta}_l = R_l \textcolor{BlueViolet}{T_l} \hat{D}_l \textcolor{BlueViolet}{T_{\rm prev}}^T R^T_{\rm prev}$. 
\\
\indent Let us first look at the $\hat{D}_l \textcolor{BlueViolet}{T_{\rm prev}}^T$ operation. $\textcolor{BlueViolet}{T_{\rm prev}}^T$ is a block diagonal matrix constructed as defined in section \ref{sec:tff_construction}. It has $\rho$ blocks along the diagonal, each with $k$ rows and at most $\lceil \frac{d}{\rho} \rceil + 2$ columns. The order of the computations required to generate this matrix is $\mathcal{O}(dk)$. The computation complexity of $\hat{D}_l \textcolor{BlueViolet}{T_{\rm prev}}^T$ is $\mathcal{O}(\frac{d}{\rho}k\rho dr) = \mathcal{O}(d^2 kr)$. So, the overall computational complexity for the computation of   $\textcolor{BlueViolet}{T_{\rm prev}}^T$ and multiplication with $\hat{D}_l$ is $\mathcal{O}(d^2 kr)$.
\\
\indent Now, consider the left multiplication with $\textcolor{BlueViolet}{T_l}$. $\textcolor{BlueViolet}{T_l}$ is again a block diagonal matrix similar to $\textcolor{BlueViolet}{T_{\rm prev}}^T$. But it is multiplying a quantity with dimensions $k\rho \times d$. Hence this multiplication has a computational complexity of $\mathcal{O}(d^2 k)$. The worst-case computational complexity of multiplication with the TFF orthonormal basis of current and previous layers is $\mathcal{O}(d^2 kr)$.
\\
The final $R_l, R_{\rm prev}^T$ are orthogonal rotation matrices which can be efficiently computed in $\mathcal{O}(d^2 \log{d})$ time using random projections such as \cite{le2013fastfood} or any other efficient implementation. Combining all these calculations, the overall computational complexity of transforming the weights during inference is $\mathcal{O}(d^2 (kr + \log{d}))$. Note that since all of these are matrix operations, they run on GPU in a vectorized manner. Table \ref{tab:inference speeds} shows the inference speeds of the quantized models on a Nvidia A100 GPU.

\comment{
\item {\em Computations during Inference:} Since the Tight Fusion Frames generated by the algorithm described in section \ref{sec:tff_construction} has a block diagonal structure, transforming the quantized weights $\hat{D}_l$ from FF representation space to the regular weight space can be done efficiently in $\mathcal{O}(kd^2r)$. This can be small at low redundancy values and a small number of subspaces. Since we are also rotating the Fusion Frames, we need an additional computational computation of $\mathcal{O}(d^2 \sqrt{d})$ or $\mathcal{O}(d^2 \log (d))$ based on the choice of implementation of random orthogonal matrices. Note that any quantization scheme in the low-bit regime will incur a cost of $\mathcal{O}(d^2)$ to transform the quantized weights by scaling and shifting them. The sparse properties of the TFF construction and the efficient choice of random rotations make FrameQuant a great choice to trade off computational complexity and storage benefits while providing additional flexibility in terms of fractional bit widths. The details of these computations are in the Appendix \ref{Computational complexity calculation}.
}

\comment{
talk about how the computational complexity is linear in r
}

\end{document}